\documentclass[a4paper,fleqn]{cas-sc}

\usepackage{natbib}
\usepackage{enumitem}
\usepackage{multirow}
\usepackage{multicol}
\usepackage{hhline}
\usepackage{longtable}
\usepackage{soul}
\usepackage{adjustbox}
\usepackage{float}
\usepackage{numprint}
\npthousandsep{,}

\usepackage{graphicx}
\usepackage{subfig}
\usepackage{array, makecell}
\usepackage{verbatim}
\usepackage[utf8]{inputenc}
\usepackage[T1]{fontenc}
\usepackage[ngerman,english]{babel}
\usepackage{tabularx}  % for 'tabularx' environment and 'X' column type
\usepackage{ragged2e}  % for '\RaggedRight' macro (allows hyphenation)
\newcolumntype{Y}{>{\RaggedRight\arraybackslash}X} 
\usepackage{booktabs}  % for \toprule, \midrule, and \bottomrule macros 
\usepackage{tablefootnote}

\usepackage{subcaption}

\usepackage{bbding}
\usepackage{fontawesome}
\usepackage{xltabular}
\usepackage{ulem}

\sethlcolor{red!6} % light red background

\def\tsc#1{\csdef{#1}{\textsc{\lowercase{#1}}\xspace}}
\tsc{WGM}
\tsc{QE}
\definecolor{darkspringgreen}{rgb}{0.09, 0.45, 0.27}

\begin{document}
\let\WriteBookmarks\relax
\def\floatpagepagefraction{1}
\def\textpagefraction{.001}

% Short title
\shorttitle{The Third VoicePrivacy  Challenge: Preserving Emotional Expressiveness and Linguistic Content in Voice Anonymization}    

% Short author
\shortauthors{...}  

% Main title of the paper
\title [mode = title]{The Third VoicePrivacy  Challenge: Preserving Emotional Expressiveness and Linguistic Content in Voice Anonymization}

\author[1]{Natalia Tomashenko}[orcid=0000-0002-7125-2382]%[<options>]

% Corresponding author indication
\cormark[1]

% Footnote of the first author
% \fnmark[1]

% Email id of the first author
\ead{natalia.tomashenko@inria.fr}

% URL of the first author
\ead[url]{https://sites.google.com/view/natalia-tomashenko}

\affiliation[1]{organization={Universit\'e de Lorraine, CNRS, Inria, LORIA, F-54000},
            % addressline={Universit\'e de Lorraine, CNRS, Inria, LORIA, F-54000}, 
            city={Nancy},
            country={France}}

\author[2]{Xiaoxiao Miao}[orcid=0000-0002-6645-6524]%[]

% Footnote of the second author
% \fnmark[2]

% Email id of the second author
\ead{xiaoxiao.miao@dukekunshan.edu.cn}

% URL of the second author
\ead[url]{https://xiaoxiaomiao323.github.io/}

\affiliation[2]{organization={Duke Kunshan University},
            % addressline={}, 
            city={Kunshan},
%          citysep={}, % Uncomment if no comma needed between city and postcode
            postcode={215316}, 
            % state={},
            country={China}}

% 3 ------------------------------------
\author[3]{Pierre Champion}[orcid=0000-0003-1225-7957]%[]
\ead{pierre.champion@inria.fr}
% \ead[url]{orcid=0000-0003-1225-7957}
\affiliation[3]{organization={Mission Défense et Sécurité, Inria},
            % addressline={}, 
            city={Grenoble},
            country={France}}

\author[4]{Sarina Meyer}[orcid=0009-0004-1117-4783]%[]
\ead{sarina.meyer@ims.uni-stuttgart.de}
% \ead[url]{orcid=0009-0004-1117-4783}
\affiliation[4]{organization={Institute for Natural Language Processing, University of Stuttgart},
            country={Germany}}
% 4 ------------------------------------
\author[5]{Michele Panariello}[orcid=0009-0007-4154-5460]%[]
\ead{michele.panariello@eurecom.fr}
\affiliation[5]
{organization={Audio Security and Privacy Group, EURECOM}, 
            city={Biot, Sophia Antipolis},
            postcode={F-06904}, 
            % state={},
            country={France}}
% 7 ------------------------------------
% 5 ------------------------------------
\author[6]{Xin Wang}[orcid=0000-0001-8246-0606]%[]
\ead{wangxin@nii.ac.jp}
\affiliation[6]{organization={National Institute of Informatics},
             addressline={Chiyoda-ku}, 
            city={Tokyo},
            postcode={101-8340}, 
            % state={},
            country={Japan}}

\author[5]{Nicholas Evans}[orcid=0000-0002-8459-1041]%[]
\ead{evans@eurecom.fr}
\ead[url]{https://www.eurecom.fr/en/people/evans-nicholas}

\author[1]{Emmanuel Vincent}[orcid=0000-0002-0183-7289]%[]
\ead{emmanuel.vincent@inria.fr}
\ead[url]{https://members.loria.fr/EVincent/}

\author[6]{Junichi Yamagishi}[orcid=0000-0003-2752-3955]
\ead{jyamagis@nii.ac.jp}
\ead[url]{https://yamagishilab.jp/}

% 10 ------------------------------------
\author[5]{Massimiliano Todisco}[orcid=0000-0003-2883-0324]
\ead{massimiliano.todisco@eurecom.fr}
\ead[url]{https://www.eurecom.fr/en/people/todisco-massimiliano}

\cortext[1]{Corresponding author}

\begin{abstract}
We present results and analyses from the third VoicePrivacy Challenge held in 2024, which focuses on advancing voice anonymization technologies. 
The task was to develop a voice anonymization system for speech data that conceals a speaker's voice identity while preserving linguistic content and emotional state. 
We provide a systematic overview of the challenge framework, including detailed descriptions of the anonymization task and datasets used for both system development and evaluation. 
We outline the attack model and objective evaluation metrics for assessing privacy protection (concealing speaker voice identity) and utility (content and emotional state preservation). We describe six baseline anonymization systems and summarize the innovative approaches developed by challenge participants. Finally, we provide key insights and observations to guide the design of future VoicePrivacy challenges and identify promising directions for voice anonymization research.

\end{abstract}


\maketitle

{\section{Introduction}\label{sec:intro}}

Speech data fall within the scope of privacy regulations such as the European General Data Protection Regulation (GDPR). 
This is because recordings of speech capture a wealth of personal and sensitive information such as the speaker's identity, age and gender, health status, personality, racial or ethnic origin, geographical background, social identity, and socio-economic status \citep{Nautsch-PreservingPrivacySpeech-CSL-2019}.
The VoicePrivacy initiative and challenge series \citep{tomashenko2020introducing} were formed in 2020 to spearhead a community effort to develop privacy preservation solutions for speech technology. 

Privacy preservation requires a combination of complementary solutions to obfuscate not only the speaker's identity, but also specific linguistic content, extra-linguistic traits, and background sounds which
might also reveal the speaker's identity or other sensitive information.
Thus far, the focus in VoicePrivacy has been upon the development of {\it voice anonymization} solutions which disguise the voice identity while preserving a set of additional attributes related to one or more downstream task (e.g. automatic speech recognition).
Common datasets and experimental protocols have been identified and designed to support the fair benchmarking of competing anonymization solutions using a set of evaluation metrics.
The first two editions of VoicePrivacy were held in 2020 \citep{tomashenko2020introducing,Tomashenko2021CSl,Tomashenko2021CSlsupplementay, tomashenkovoiceprivacy,results2022} and 2022 \citep{tomashenko2022voiceprivacy,results2022,VPC2024}, resulting in substantial progress and the forming of a new research community in voice anonymization.

In this paper we describe the most recent edition of the VoicePrivacy Challenge (VPC) which concluded in~2024.
In keeping with previous editions,
the 2024 edition maintained the focus on voice anonymization and the preservation of linguistic content and paralinguistic attributes.
New to the 2024 edition was a specific requirement to preserve the speaker's emotional state, a  key paralinguistic attribute in many real-world application scenarios for voice anonymization, e.g., in call centers to enable the use of third-party speech analytics.\footnote{This requirement reflects a selective anonymization approach: while speaker identity cues are suppressed to ensure GDPR compliance, certain paralinguistic attributes such as emotion are deliberately preserved to support downstream utility. Preserving emotion is generally considered compatible with GDPR provided that it cannot be linked back to the individual or reveal sensitive health information.}

Placing emphasis on preserving emotional attributes in voice anonymization raises the inherent difficulty of the task; it requires concealing the voice identity while maintaining an increasingly rich set of linguistic and paralinguistic characteristics. 
Successive challenge editions should therefore aim to broaden -- and even shift -- their scope, with a more explicit focus not only on what must be suppressed or obfuscated, but equally on what should be preserved after anonymization. 
Introducing such balanced and progressively demanding constraints can serve as a mechanism to stimulate scientific progress, advancing speech representation disentanglement, and promoting more robust and principled approaches to voice anonymization.
The new focus also demands the adoption of additional emotion-labeled databases and experimental protocols, as well as new metrics to support the evaluation of emotion preservation.

The differences among the three VoicePrivacy Challenge editions are summarized in Table~\ref{tab:vpc_compare}. It highlights the progressive evolution of objectives, attacker strength, evaluation strategies, and methodological diversity from 2020 to 2024:
VPC 2020 focused on 
concealing speaker voice identity (under various attack scenarios: \textit{ignorant}, \textit{lazy-informed}, \textit{semi-informed}) while preserving linguistic content, naturalness, and voice distinctiveness as a complementary objective. 
% 2022
VPC 2022 strengthened defenses against 
\textit{semi-informed} attackers and introduced intonation preservation; 
% 2024
and VPC 2024 extended requirements to preserve 
both linguistic and emotional expressiveness. 
This progression reflects increasingly advanced attacker models, an expanding range of evaluation metrics from basic utility measures to prosody  and emotion preservation,
and significant growth in  diversity of anonymization methods  ---  with submitted systems evolving from incremental improvements over  baselines to using neural codecs, advanced attribute disentanglement methods, and hybrid approaches that flexibly balance privacy-utility tradeoffs. These developments underscore the field's advancing demands for robust, multi-dimensional voice privacy solutions.

\begin{table*}[t]
\centering
\caption{Comparison of VoicePrivacy Challenges (2020–2024)}
\label{tab:vpc_compare}
\begin{adjustbox}{max width=\linewidth}
\begin{tabular}{lll}
\toprule
\textbf{VoicePrivacy 2020} & \textbf{VoicePrivacy 2022} & \textbf{VoicePrivacy 2024} \\
\midrule
\multicolumn{3}{l}
{\cellcolor{gray!25}\textbf{Task}} \\
\makecell[l]{remove speaker id \\ keep linguistic content \\ keep  naturalness $\&$ intelligibility}
&
\makecell[l]{remove speaker id \\ keep linguistic content \\ \textbf{keep intonation} \\  keep  naturalness $\&$ intelligibility}
&
\makecell[l]{remove speaker id \\ keep linguistic content \\ \textbf{keep emotional state}}
\\
\midrule
\multicolumn{3}{l}{\cellcolor{gray!25}\textbf{Training data}} \\
\makecell[l]{LibriSpeech: train-clean-100, train-other-500 \\ 
LibriTTS: train-clean-100, train-other-500 \\ 
VoxCeleb-1,2} &
\makecell[l]{LibriSpeech: train-clean-100, train-other-500 \\ 
LibriTTS: train-clean-100, train-other-500 \\ 
VoxCeleb-1,2} &
\makecell[l]{\textbf{open resources (data, models)} \\ \textbf{(based on participant proposals)}} \\

\midrule
\multicolumn{3}{l}{\cellcolor{gray!25}\textbf{Development and evaluation data}} \\
\makecell[l]{LibriSpeech dev-/test-clean \\ VCTK} &
\makecell[l]{LibriSpeech dev-/test-clean \\ VCTK} &
\makecell[l]{LibriSpeech dev-/test-clean \\ \textbf{IEMOCAP}} \\

%level
\midrule
\multicolumn{3}{l}{\cellcolor{gray!25}\textbf{Anonymization level of the test data
}} \\
\makecell[l]{speaker} &
speaker &
\textbf{utterance} \\ 
\midrule
\multicolumn{3}{l}
{\cellcolor{gray!25}\textbf{Objective assessment}} \\
\multicolumn{3}{l}{\cellcolor{Gray!5}\textbf{Attack models}} \\
\textit{ignorant}, \textit{lazy-informed}, \textit{semi-informed} &
\textbf{\textit{semi-informed}} &
\textit{semi-informed} \\ 
\multicolumn{3}{l}{\cellcolor{Gray!5}\textbf{Privacy metrics}} \\
\makecell[l]{Primary: equal error rate (EER), \\
Post-eval: $C_\text{llr}$, $C_\text{llr}^\text{min}$, linkability, \\ ZEBRA~\citep{nautsch2020zebra}, \\ 
de-identification (DeID)~\citep{noe2020speech,noe2021csl}} &
EER &
EER \\ 
\multicolumn{3}{l}{\cellcolor{Gray!5}\textbf{Utility metrics}} \\
\makecell[l]{WER (ASR trained on orig. \& anon. data) \\ 
Gain of voice distinctiveness ($G_\text{VD}$)} &
\makecell[l]{WER (ASR trained on \textbf{anonymized} data) \\ Gain of voice distinctiveness ($G_\text{VD}$) \\ \textbf{Pitch correlation} ($\rho_{F0} > 0.3$)}  &
\makecell[l]{WER (ASR trained on \textbf{original} data)  \\
\textbf{UAR (SER trained on original data)}} \\ 
\midrule
\multicolumn{3}{l}{\cellcolor{gray!25}\textbf{Subjective assessment}} \\
\multicolumn{3}{l}{\cellcolor{Gray!5}\textbf{Privacy metrics}} \\
\makecell[l]{Primary: speaker verifiability \\
Post-eval: linkability~\citep{o2021anonymous}} &
speaker verifiability &
--- \\ 
\multicolumn{3}{l}{\cellcolor{Gray!5}\textbf{Utility metrics}} \\
\makecell[l]{naturalness \& intelligibility} &
\makecell[l]{naturalness \& intelligibility} &
--- \\ 
%
%
%
%ranking
\midrule
\multicolumn{3}{l}{\cellcolor{gray!25}\textbf{Ranking policy
}} \\
\makecell[l]{---} &
\makecell[l]{\textbf{ranking by utility  within  privacy} \\ \textbf{categories}    (\textbf{EER}  $\geq$ \textbf{15\%, 20\%, 25\%, 30\%})} &
\makecell[l]{ranking by utility within  privacy \\ categories  (\textbf{EER } $\geq$ \textbf{10\%, 20\%, 30\%, 40\%})}\\ 
\midrule
\multicolumn{3}{l}
{\cellcolor{gray!25}\textbf{Baselines}} \\

\makecell[l]{
\textit{-- Neural voice conversion:} \\B1 (x-vector + neural vocoder) \\ \textit{-- Signal processing:} \\ B2 (McAdams fixed coeff.)} & \makecell[l]{\textit{-- Neural voice conversion:} \\ \textbf{B1.a, B1.b} \\
 \textit{-- Signal processing:} \\ B2 (McAdams \textbf{random} coeff.)} & \makecell[l]{B1 (B1.b from VPC2022) \\
B2 (McAdams random coeff.) \\
\textbf{B3 (phonetic transcriptions + TTS and GAN)}\\
\textbf{B4 (neural codecs)} \\
\textbf{B5, B6 (ASR-BN with VQ)}} \\

\midrule
\multicolumn{3}{l}{\cellcolor{gray!25}\textbf{Submitted systems}} \\
\makecell[l]{\textit{-- Neural voice conversion:} \\ 
Attribute disentanglement \\ (focus on x-vector anonymization, \\ mainly incremental improvements \\ over the baselines) \\ 
\textit{-- Signal processing:} \\ 
Formant, F0, and speaking rate modification} &
\makecell[l]{\textit{-- Neural voice conversion:} \\ 
Attribute disentanglement \\ (\textbf{modify all components in baseline}: \\
content,  speaker,  and
prosody \\ feature extraction;  speaker and prosody \\ anonymizers;    speech generation models) \\ 
\textit{-- Signal processing:} \\ 
Formant, pitch, and speaking rate modification} &
\makecell[l]{\textit{-- Neural voice conversion:} \\
Attribute disentanglement \\ (speaker, linguistic, pitch, \textbf{emotion}; \\\textbf{ multiple emotion preservation strategies} \\(\textbf{pretrained encoders}, \textbf{distillation}); \textbf{kNN-VC} \\ 
\textit{-- \textbf{Neural codecs}} \\
\textit{-- \textbf{ASR+TTS cascades}} \\ 
\textit{-- \textbf{Hybrid methods}}} \\
\bottomrule
\end{tabular}
\end{adjustbox}
\end{table*}

In this paper we describe the VPC 2024, present the challenge task, attack model, protocols, anonymization baselines and evaluation metrics.
Furthermore, we provide a summary of the 36 anonymization systems designed by challenge participants, extending the analyses presented at the VoicePrivacy 2024 workshop held in conjunction with the 4th Symposium on Security and Privacy in Speech Communication (SPSC),\footnote{\url{https://spsc-symposium.de/2024}} a joint event co-located with Interspeech 2024.
We present new comparisons to systems that emerged from previous VPC editions, with additional insights into limitations and the reliability of privacy evaluation in the face of stronger attacks.
We conclude with an overview of post-challenge results, and our perspectives on future research and likely directions for the next VPC edition tentatively scheduled for 2026.

{\section{Challenge design}}\label{sec:challenge_design}

In this section, we present an overview of the challenge setup: the anonymization task,  attack model, and data.

\subsection{Voice anonymization task}
\label{subsec:task}

\begin{figure}[!htbp]
\centering\includegraphics[width=0.98\textwidth]{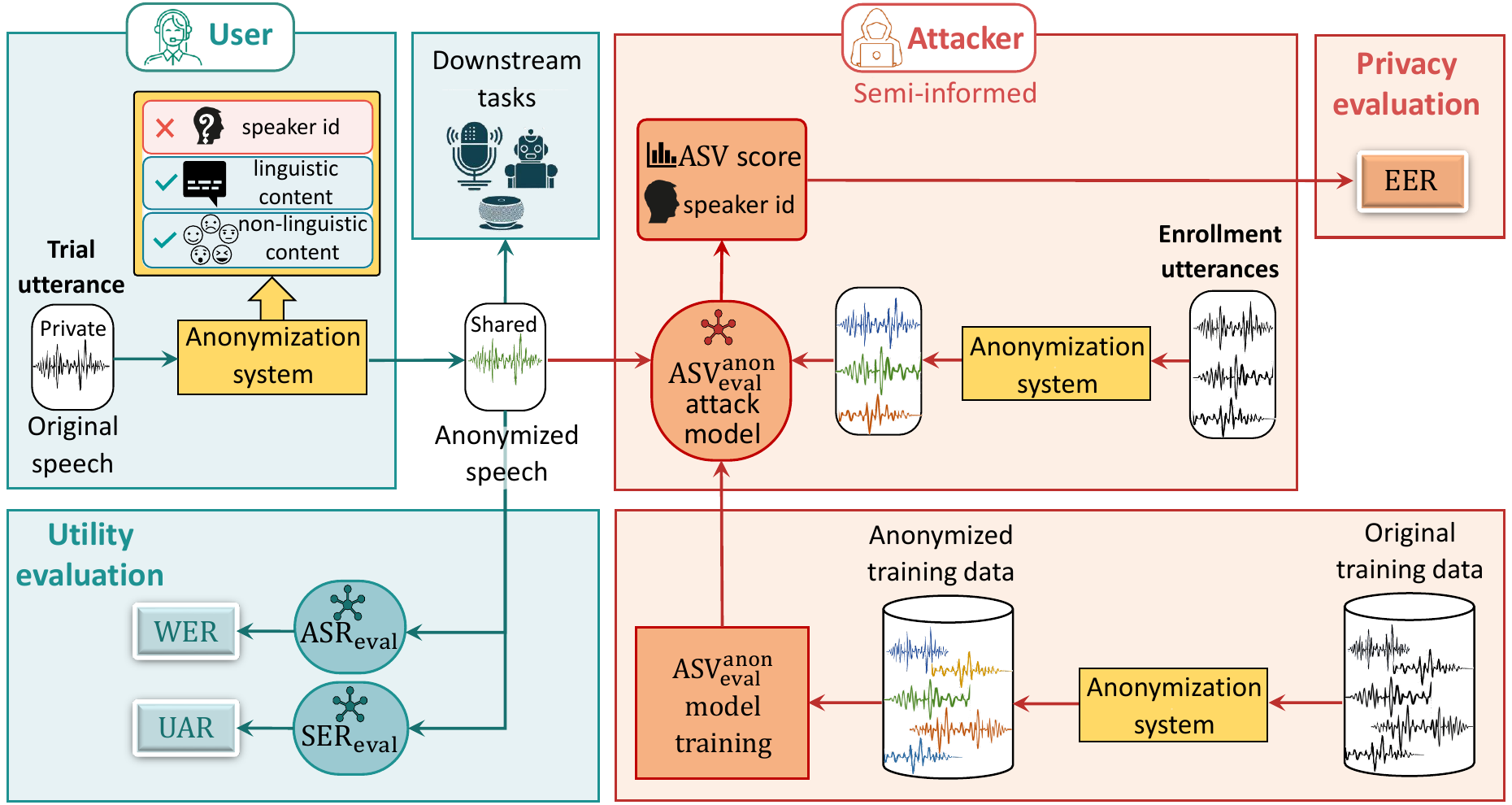}
\caption{
Privacy preservation scenario as a game between \textit{users} and \textit{attackers} in the case where speaker identity is considered as personal information to be protected using anonymization, while linguistic and emotional content should be preserved for utility downstream tasks. 
Privacy evaluation of anonymized data is performed using a \textit{semi-informed} attack model   $ASV_\text{eval}^{\text{anon}}$  and equal error rate (EER) metric. Utility evaluation is performed using (1) automatic speech recognition (ASR) model $ASR_\text{eval}$ and 
word error rate (WER) metric and 
(2) speech emotion recognition (SER) model 
$SER_\text{eval}$ 
and unweighted average recall (UAR) metric. 
}
\label{fig:privacy-task}
\end{figure}

Privacy protection is formulated as a game between a \textit{user} who 
shares data with the purpose of accomplishing some desired downstream task and an \textit{attacker} who accesses and uses this data, or data derived from it, to infer information about the data subject~\citep{qian2018towards,srivastava2019evaluating,tomashenko2020introducing}. 
Here, we consider the scenario where the user shares anonymized utterances for downstream automatic speech recognition (ASR) and speech emotion recognition (SER) tasks, whereas the attacker seeks to identify the speaker using the same anonymized utterances (see Figure~\ref{fig:privacy-task}).

%%%%%%%%%%%%%%%%%
Utterances shared by the user are referred to as \textit{trial} utterances. 
In order to hide his or her identity, the user treats each utterance with a voice anonymization system prior to sharing. 
The resulting utterance sounds as if it was uttered by another speaker, referred to as a \textit{pseudo-speaker}. 
The pseudo-speaker might, for instance, be an artificial voice not corresponding to the voice of any real individual.
The task of challenge participants is to develop this voice anonymization system which should (a)~output a speech waveform; (b)~conceal the voice identity at the \textit{utterance level}; 
(c) preserve the linguistic and emotional content.

Requirement (b) for \textit{utterance-level} anonymization means that the voice anonymization system must replace the voice identity with that of a pseudo-speaker independently of any other utterance. 
The pseudo-speaker assignment process must be identical across all utterances and not rely on speaker labels. 
When this process involves use of a random number generator, the random number(s) generated for each utterance must be different, thereby resulting in a different pseudo-speaker for each utterance. Alternatively, a single pseudo-speaker may be assigned to {\it all} utterances to satisfy this requirement.
The achievement of requirement (c) is assessed via \textit{utility} metrics.
Specifically, we  measure the preservation of linguistic and emotional content using the word error rate (WER) and unweighted average recall (UAR) using automatic speech recognition (ASR) and speech emotion recognition (SER) systems, both trained using original (unprotected) data.

\subsection{Attack model}
\label{subsec:attack_model}

For each speaker of interest, the attacker is assumed to gain access to utterances spoken by that speaker, referred to as \textit{enrollment} utterances. 
Using these utterances, attackers 
% can
deploy automatic speaker verification (ASV) to re-identify the speaker whose voice is contained in each anonymized \textit{trial} utterance.

We assume that the attacker has access to:
% \begin{enumerate}[label=(\alph*)]
(a) several enrollment utterances for each speaker
and (b) the same voice anonymization system employed by the user.
% \end{enumerate}
Using these resources, 
% ....
the attacker anonymizes the enrollment 
utterances to reduce the mismatch with the trial utterances, and trains an ASV system adapted to that specific anonymization system. 
Like the user, the attacker performs voice anonymization at the \textit{utterance level}.
% All the utterances for the attack model training are anonymized at the \textit{utterance level}.
% This attack model is the strongest known to date, hence we consider it as the most reliable for privacy assessment.
This attack model was, to date, conceptually the strongest known; therefore, we considered it 
% the most reliable 
for privacy assessment.
%
% The protection of identity information 
% % anonymization performance 
% is assessed using 
%  a \textit{privacy} metric. Specifically, we will measure 
% the ASV EER.  The better the anonymization, the more challenging it should be to re-identify speakers and, hence, the higher the EER.
% %obtained by the attacker.

\subsection{Data and pretrained models}\label{sec:data}
%Publicly available resources are used for 
The training, development and evaluation of voice anonymization systems are all performed using publicly available resources. The development and evaluation data are fixed, while the choice of training resources is flexible, subject to the constraints outlined below.
%and left open to the participants.

\subsubsection{Training resources}
In addition to the training data used for previous challenge editions (\textit{VoxCeleb-1,2}; 
% \citep{nagrani2017voxceleb,chung2018voxceleb2}, 
\textit{LibriSpeech train-clean-100}, \textit{train-other-500}; 
% \citep{panayotov2015librispeech},
\textit{LibriTTS train-clean-100}, 
\textit{train-other-500}) 
% \citep{zen2019libritts}) 
as well as data and pretrained models used to build and develop baseline anonymization systems (corpora \textit{RAVDESS},
% \citep{livingstone2018ryerson}, 
\textit{ESD}; 
% \citep{zhou2021seen}
 models \textit{HuBERT Base}, \textit{EnCodec}, 
% \citep{encodec}, 
\href{https://huggingface.co/suno/bark}{\textit{Bark}}, and \href{https://dl.fbaipublicfiles.com/voxpopuli/models/wav2vec2_large_west_germanic_v2.pt}{\textit{wav2vec2\_large\_west\_germanic\_v2}}) ---  see Section~\ref{sec:baseline} for more details --- participants were permitted to propose additional training resources, 
%to build and train anonymization systems before the specified deadline.  These 
including both data and pretrained models.
From the suggestions received, a list of permitted training data and pretrained models was  finalized and published in an update to the evaluation plan~\citep{tomashenko2024voiceprivacy}. The list is shown in Table~\ref{tab:data-models-final-list} in Appendix~\ref{app:trainig_res}.

\subsubsection{Development and evaluation data}\label{subsec:dev_eval_data}

Development and evaluation datasets
come from two corpora:
\textit{LibriSpeech}\footnote{\label{fn:url1}LibriSpeech: \url{http://www.openslr.org/12}} \citep{panayotov2015librispeech} and \textit{IEMOCAP} \citep{Busso08}, as presented in Tables~\ref{tab:data} and \ref{tab:data_imo}.
\textit{LibriSpeech}  contains 960 hours of read English speech (16 kHz) in the form of audiobooks and is used commonly for ASV and ASR evaluation.
The \textit{LibriSpeech} evaluation and development sets are the same as for previous challenge editions. 
%
%From the \textit{LibriSpeech} corpus, official test
%\textsuperscript{\ref{fn:url1}}
%
    % \item \textit{\textbf{LibriTTS}}\footnote{LibriTTS: \url{http://www.openslr.org/60/}} \citep{zen2019libritts} is a corpus of English speech derived from LibriSpeech and designed for research in text-to-speech (TTS). It contains approximately 585~hours of read English speech sampled at 24~kHz.
%
    % \item {\textbf{\textit{VCTK}}}\footnote{VCTK, release version 0.92: \url{https://datashare.is.ed.ac.uk/handle/10283/3443}} \citep{yamagishi2019cstr} is a corpus of read speech collected from 109 native speakers of English with various accents. It was originally aimed for research in TTS and contains approximately 44~hours of speech sampled at 48~kHz.
 %   
    % \item \textit{\textbf{VoxCeleb-1,2}}\footnote{VoxCeleb: \url{http://www.openslr.org/60/}} \citep{nagrani2017voxceleb,chung2018voxceleb2} is an audiovisual corpus extracted from videos uploaded to YouTube and designed for speaker verification research. It contains
    % approximately \numprint{2770}~hours of speech sampled at 16~kHz collected from \numprint{7363}~speakers, covering a wide range of accents and languages.
%
   % \item 
 %  
\textit{IEMOCAP}  
contains 12h of emotional speech (16 kHz) collected from two-speaker conversations between 10 English actors (5 female, 5 male).
%
% We consider only 4 emotions out of the 9 annotated ones: \textit{neutral}, \textit{sadness}, \textit{anger}, and \textit{happiness}.
% %
% Following \citep{Pappagari2020XVectorsME,asr_IEMOCAP,Nourtel_spsc_emotion}, we merge the original happiness and excitement classes into the happiness class to balance the number of utterances in each class.
% To accommodate for the small number of speakers and the small amount of data, we adopt a leave-one-conversation out cross-validation protocol. 
Following \citet{Pappagari2020XVectorsME}, we focus on a subset of 4 emotions for SER evaluation: \textit{neutral}, \textit{sadness}, \textit{anger}, and \textit{happiness} (with \textit{happiness} and \textit{excitement} classes merged).
Due to data limitations, we implement leave-one-conversation-out cross-validation, using four conversations (8 speakers) for SER training and the remaining one conversation (2 speakers) for development and evaluation.

\begin{table}[!htbp]
\centering
\caption{Statistics of the \textit{LibriSpeech} development and evaluation sets for ASV and ASR evaluation.}
\label{tab:data}
\begin{tabular}{cllcccr}
\toprule
 & \textbf{Subset} & \textbf{Type} & \multicolumn{3}{c}{\textbf{\#Speakers}} & \textbf{\#Utterances} \\ 
\cmidrule(lr){4-6}
 &  &  & \textbf{Female} & \textbf{Male} & \textbf{Total} &  \\ 
\midrule
\multirow{2}{*}{Development} 
 & \multirow{2}{*}{LibriSpeech dev-clean} 
    & Enrollment & 15 & 14 & 29 & 343 \\ 
 &  & Trial & 20 & 20 & 40\textsuperscript{*} & \numprint{1978} \\ 
\cmidrule{1-7}
\multirow{2}{*}{Evaluation} 
 & \multirow{2}{*}{LibriSpeech test-clean} 
    & Enrollment & 16 & 13 & 29 & 438 \\ 
 &  & Trial & 20 & 20 & 40\textsuperscript{*} & \numprint{1496} \\ 
\bottomrule
\end{tabular}

\vspace{2mm}
% \footnotesize{\textsuperscript{*}Includes 11 speakers appearing only as non-target (impostor) speakers in the trials.}
\footnotesize{\textsuperscript{*}Includes 11 speakers who appear only in \textit{different-speaker} trials.}
\end{table}

\begin{table}[!htbp]
\centering
  \caption{Construction and statistics of the \textit{IEMOCAP}  development and evaluation sets for SER evaluation. \textit{Train} subsets refer to the training data for the SER evaluation system.
  }\label{tab:data_imo}
 % \resizebox{0.95\textwidth}{!}{
  \centering
  \begin{tabular}{clcccccc}
\toprule
 \multicolumn{2}{l}{\textbf{Conversation}}  & \textbf{\#Utterances}  & \textbf{Fold 1} & \textbf{Fold 2} & \textbf{Fold 3} & \textbf{Fold 4} & \textbf{Fold 5}\\ 
\midrule

\multirow{2}{*}{\rotatebox{0}{~ Session 1 }}  &  Female  & 528 & Dev &  \multirow{2}{*}{\rotatebox{0}{{Train}}} &  \multirow{4}{*}{\rotatebox{0}{{Train}}} &  \multirow{6}{*}{\rotatebox{0}{{Train}}} &  \multirow{8}{*}{\rotatebox{0}{{Train}}} \\\cmidrule{2-4}
 & Male  & 557 & Eval &&&&\\ \cmidrule{1-5}
\multirow{2}{*}{\rotatebox{0}{~ Session 2 }}  &  Female  & 481 & \multirow{8}{*}{\rotatebox{0}{{Train}}} & Eval &&&\\ \cmidrule{2-3}\cmidrule{5-5}
 & Male  & 542 & & Dev &&&\\ \cmidrule{1-3}\cmidrule{5-5}\cmidrule{6-6}
 \multirow{2}{*}{\rotatebox{0}{~ Session 3 }}  &  Female  & 522 & & \multirow{6}{*}{\rotatebox{0}{{Train}}} & Dev&&\\ \cmidrule{2-3}\cmidrule{6-6}
 & Male  & 629&&  &Eval&&\\ \cmidrule{1-3}\cmidrule{7-7}\cmidrule{6-6}
  \multirow{2}{*}{\rotatebox{0}{~ Session 4 }}  &  Female  & 528&&  &\multirow{4}{*}{\rotatebox{0}{{Train}}} &Eval&\\ \cmidrule{2-3} \cmidrule{7-7}
 & Male  & 503&&  &&Dev&\\ \cmidrule{1-3} \cmidrule{7-7}\cmidrule{8-8}
   \multirow{2}{*}{\rotatebox{0}{~ Session 5 }}  &  Female  & 590&&  &&\multirow{2}{*}{\rotatebox{0}{{Train}}}&Eval\\ \cmidrule{2-3}\cmidrule{8-8}
 & Male  & 651&& &&&Dev\\
 % \cline{1-3}\cline{8-8}
 % \Xhline{0.7pt}
 \bottomrule
\end{tabular}
% }
\end{table}

% \textcolor{blue}{\section{Privacy and utility evaluation}}
% \label{sec:perf}

% We consider one objective privacy metric to assess the speaker re-identification risk and two objective utility metrics to assess the fulfillment of the downstream tasks specified in Section~\ref{subsec:task}. 

\section{Privacy and utility evaluation}
\label{sec:perf_objective}

Three metrics are used for the objective ranking of submitted systems: the equal error rate (EER) as the privacy metric and two utility metrics: word error rate (WER) and unweighted average recall (UAR).
These metrics rely on automatic speaker verification (ASV),  automatic speech recognition (ASR), and speech emotion recognition (SER) systems.

\subsection{Training data for evaluation models}\label{ssec:train-data-for-eval-models}

The ASV system is trained on \textit{LibriSpeech-train-clean-360} and the ASR system on the full \textit{LibriSpeech-train-960} dataset, whose statistics are presented in Table~\ref{tab:train-eval-metrics}. 
The SER system for each \textit{IEMOCAP} cross-validation fold is trained on the corresponding \textit{IEMOCAP} training subset, whose statistics are reported in Table~\ref{tab:data_imo}.
Training and evaluation are performed with the provided recipes and models.\footnote{Evaluation scripts: \url{https://github.com/Voice-Privacy-Challenge/Voice-Privacy-Challenge-2024}} 
% More specifically,
Models for privacy evaluation are trained by participants on their anonymized training data with the provided training scripts, while the models for utility evaluation are provided by the organizers.

\begin{table}[!htbp]
  \caption{Statistics of the \textit{LibriSpeech} training sets for ASV and ASR evaluation models.}\label{tab:train-eval-metrics}
  \renewcommand{\arraystretch}{1.1}
  \centering
  \begin{tabular}{cccrrrr}
\toprule
  \multirow{2}{*}{\textbf{System}} & \multirow{2}{*}{\textbf{Subset}} &   \multirow{2}{*}{\textbf{Size,h}} & \multicolumn{3}{c}{\textbf{\#Speakers}} &  \multirow{2}{*}{\textbf{\#Utterances}} \\ \cmidrule{4-6}
&  &  & \textbf{Female} & \textbf{Male} & \textbf{Total} & \\ \midrule
 %$ASV_\text{eval}^\text{anon}$ 
 ASV
 & LibriSpeech train-clean-360 & 363.6 & 439 & 482	 &	921	& \numprint{104014}	\\ \midrule
 ASR
 & LibriSpeech train-960 & 960.9 & 1128 & 1210	 &	2338	& \numprint{281241}	\\ 
 \bottomrule
  \end{tabular}
\end{table}

\subsection{Privacy metric: the equal error rate (EER)}\label{sec:asv-eval}

The ASV system used for privacy\footnote{In this context, 
\textit{privacy} refers specifically to speaker de-identification (measured by ASV EER increase), while \textit{utility} denotes content and emotion preservation (ASR WER, SER UAR); together they constitute \textit{anonymisation performance}.
More broadly, \textit{privacy} encompasses the protection of all personal and sensitive information that could identify or link an individual, including metadata, contextual details, and behavioral patterns beyond voice characteristics. Therefore, even perfect speaker de-identification may not guarantee complete privacy if other identifiable information, such as a speaker's name, location, or interaction patterns, is exposed or disclosed.
%%%
}
evaluation is an ECAPA-TDNN model~\citep{desplanques2020ecapa} with 512 channels in the convolution frame layers, implemented by adapting the \textit{SpeechBrain}~\citep{speechbrain} recipe for the \textit{VoxCeleb} dataset to the \textit{LibriSpeech} dataset.
As illustrated in Figure~\ref{fig:privacy-task}, the attacker is assumed to have access to the anonymization system under evaluation~\citep{srivastava2019evaluating,Tomashenko2021CSl}. The attacker uses the system to anonymize the enrollment utterance(s) so as to reduce the mismatch between enrollment and trial utterances.
In addition, the attacker uses the same system to anonymize the \textit{LibriSpeech-train-clean-360} dataset and retrains the ASV system (denoted $ASV_\text{eval}^{\text{anon}}$) so that it is adapted to operate upon similarly anonymized input.
In all cases, anonymization is applied at the \textit{utterance level}, using the same pseudo-speaker assignment process used for anonymization of trial utterances.  
For speakers with multiple enrollment utterances, embeddings are extracted for each utterance and then averaged.

\begin{table}[!htbp]
  \caption{Number of \textit{same-speaker} and \textit{different-speaker} pairs considered for evaluation.}\label{tab:trials}
  \centering
   % \resizebox{0.87\textwidth}{!}
   {
  \begin{tabular}{lllrrr}
\toprule
 \multicolumn{2}{l}{\textbf{Subset}} & \textbf{Trials} &  \textbf{Female} & \textbf{Male} & \textbf{Total}  \\ \midrule
% dev
\multirow{2}{*}{\rotatebox{0}{Development~}} & \multirow{2}{*}{LibriSpeech dev-clean} & Same-speaker & 704 & 644 & \numprint{1348} \\ \cmidrule{3-6}
 &  & Different-speaker	& \numprint{14566} & \numprint{12796} &	\numprint{27362} \\ \cmidrule{1-6}
\multirow{2}{*}{\rotatebox{0}{Evaluation~}} & \multirow{2}{*}{LibriSpeech test-clean} & Same-speaker & 548 & 449	& \numprint{997} \\ \cmidrule{3-6}
  &  & Different-speaker & \numprint{11196} & \numprint{9457} &	\numprint{20653} \\ \bottomrule
  \end{tabular}}
\end{table}
\normalsize

The cosine similarity score is computed for each pair of embeddings extracted from enrollment and trial utterances and thresholded to produce same-speaker vs.\ different-speaker decisions.
Denoting the ASV false alarm and miss rates at threshold~$\theta$ by $P_\text{fa}(\theta)$ and $P_\text{miss}(\theta)$ respectively, the EER  
is defined by the decision threshold $\theta_\text{EER}$ for which the two detection error rates are equal, i.e., $\text{EER}=P_\text{fa}(\theta_\text{EER})=P_\text{miss}(\theta_\text{EER})$. The higher the EER, the better the anonymization.
The number of same-speaker and different-speaker enrollment/trial pairs used to estimate the EER is given in Table~\ref{tab:trials} for the development and evaluation partitions of each dataset.

\subsection{Utility metrics}\label{sec:utility_metrics}

Whereas for the 2022 VPC there were three utility metrics, there are only two for the 2024 edition.  While the measure of linguistic content preservation remains unchanged, measures of pitch correlation and gain of voice distinctiveness were replaced with a measure of emotion preservation.  Measures of pitch correlation were deemed to be insufficiently informative or reliable, while the gain of voice distinctiveness pertains to an unnecessarily narrow use case scenario.  A speaker's emotional state is a key paralinguistic attribute that is relevant to a broader range of application scenarios.

\subsubsection{Word error rate (WER)}\label{sec:wer}

The preservation of linguistic content is assessed using an ASR system\footnote{\url{https://huggingface.co/speechbrain/asr-wav2vec2-librispeech}} (denoted $ASR_\text{eval}$) fine-tuned using the \textit{LibriSpeech-train-960} dataset and \textit{wav2vec2-large-960h-lv60-self}\footnote{\url{https://huggingface.co/facebook/wav2vec2-large-960h-lv60-self}} representations
using a \textit{SpeechBrain} recipe.
The ASR evaluation model is trained 
% and fine-tuned 
using original (unprocessed) data.
In contrast to previous VPC editions, however, it is then fixed rather than being adapted to anonymized data.  

The motivation for using original  data to train the ASR for utility estimation is to 
 minimize acoustic distortions and better approximate the natural speech distribution, enabling utility estimates that reflect real-world usability and listener experience with anonymized speech. 
The broader perspective involves using anonymized data for training or fine-tuning downstream models that can perform well on both original and anonymized speech.

For every utterance in the \textit{LibriSpeech} development and evaluation sets, the ASR system outputs a word sequence. The WER, estimated using either original or anonymized data, is calculated according to
\begin{equation*}
\text{WER}=\frac{N_\text{sub}+N_\text{del}+N_\text{ins}}{N_\text{ref}},
\end{equation*}
where $N_\text{sub}$, $N_\text{del}$, and $N_\text{ins}$ are the number of substitution, deletion, and insertion errors, respectively, and $N_\text{ref}$ is the number of words in the reference. %
The lower the WER, the better the preservation of linguistic content, and hence the greater the utility.

\subsubsection{Unweighted average recall (UAR)}
The preservation of emotional content is assessed using an SER system (denoted $SER_\text{eval}$) trained using the \textit{SpeechBrain} recipe for the \textit{IEMOCAP} database. The wav2vec2-based model
is trained separately for each of the training folds shown in Table~\ref{tab:data_imo}.
SER performance is measured  using the \textit{IEMOCAP} development and evaluation sets using either original or anonymized utterances.
We calculate  
the unweighted average recall (UAR) according to
\begin{equation*}
    \label{eq_uar}
   \text{UAR} = \frac{\sum_{i=1}^{N_\text{class}}{R_{i}}}{N_\text{class}}.
\end{equation*}
where the recall $R_i$ for each class $i$ is computed as the number of true positives divided by the total number of samples in that class and where $N_\text{class}=4$ is the number of classes.
UARs are averaged across the five folds.
The higher the UAR, the better the preservation of emotional state and the greater the utility.
Unweighted averaging (in UAR) rather than weighted averaging is used to prevent high-frequency emotion classes from dominating the metric, which is particularly important given class imbalance in \textit{IEMOCAP}.

\subsection{Privacy-utility tradeoff}
\label{sec:perf}
\begin{figure}[!htbp]
\centering\includegraphics[width=165mm]{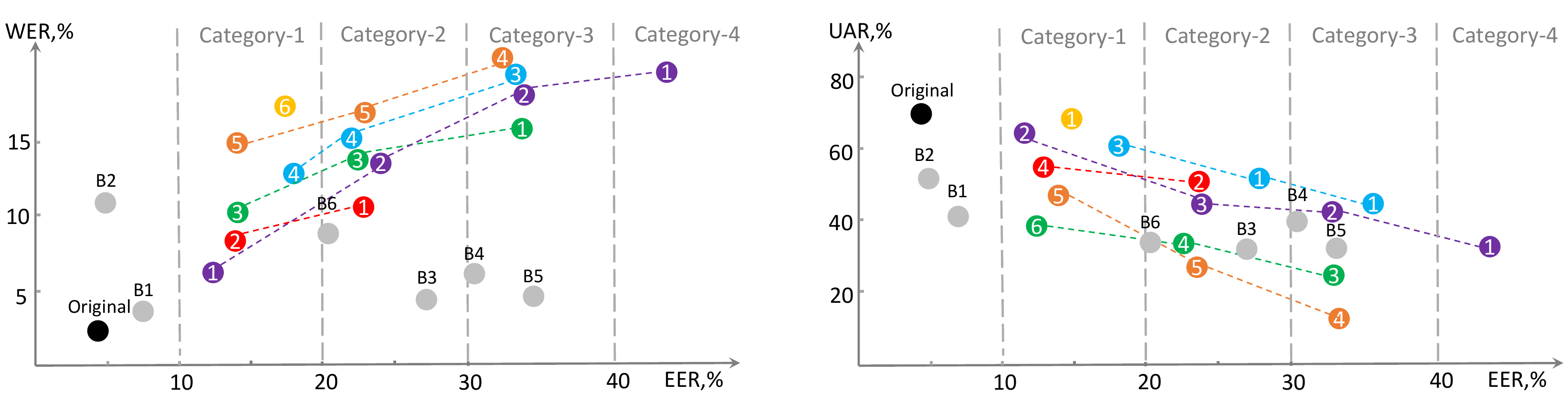}

\caption{Example system rankings according to the privacy (EER) and utility (WER and UAR) results for 4 minimum target EERs. Different colors correspond to 6 different teams. Numbers within each circle show system ranks for a given category. Grey circles correspond to the baseline systems, and the black circle to the original (unprocessed) data.}
\label{fig:thresholds}
\end{figure}

As in the 2022 edition, multiple evaluation conditions 
(privacy categories)
specified with a set of minimum target privacy requirements are considered.
For each minimum target privacy requirement, submissions that meet this requirement are ranked according to the resulting utility for each utility metric separately. 
The goal is to measure the privacy-utility trade-off at multiple operating points, e.g. when systems are configured to offer better privacy at the cost of utility and vice versa. 
This evaluation framework better aligns with real-world user privacy expectations and provides participants with clear optimization targets, creating a more comprehensive assessment using our established privacy and utility metrics.
Privacy requirements  
are defined by $N$ minimum target EERs: \{EER$_1$, \ldots, EER$_N$\}. 
Each minimum target EER constitutes a separate evaluation category.
Submissions to category $i$ had to achieve an average EER on the VoicePrivacy 2024 evaluation set greater than EER$_i$. 
The challenge includes $N=4$ categories with minimum target EERs of $10\%$, $20\%$, $30\%$, and $40\%$. Systems are ranked by WER (lower is better) and UAR (higher is better) within each  privacy (EER) category. A depiction of example results and system rankings according to this methodology is shown in Figure~\ref{fig:thresholds}.

{\section{Baseline voice anonymization systems}}\label{sec:baseline} 

In this section we provide an overview of the baseline voice anonymization systems which were made available to participants as inspiration for developing their own solutions. To gauge progress, we included two baseline systems (\textbf{B1} and \textbf{B2}) established with previous challenge editions. 
We also made four new baseline systems available, namely (\textbf{B3}, \textbf{B4}, \textbf{B5}, and \textbf{B6}), which offer enhanced anonymization performance with varying compromises to utility. 
New baseline systems are trained using different resources.
A description of each baseline is provided in the following. 
Further details can be found in the evaluation plan \citep{tomashenko2024voiceprivacy} and in the publicly available baseline recipe scripts\footnote{The VoicePrivacy 2024 Challenge anonymization baselines and evaluation scripts: \url{https://github.com/Voice-Privacy-Challenge/Voice-Privacy-Challenge-2024}}.

\subsection{Anonymization using x-vectors and a neural source-filter model: B1}

\begin{figure}[h!]
\centering\includegraphics[width=100mm]{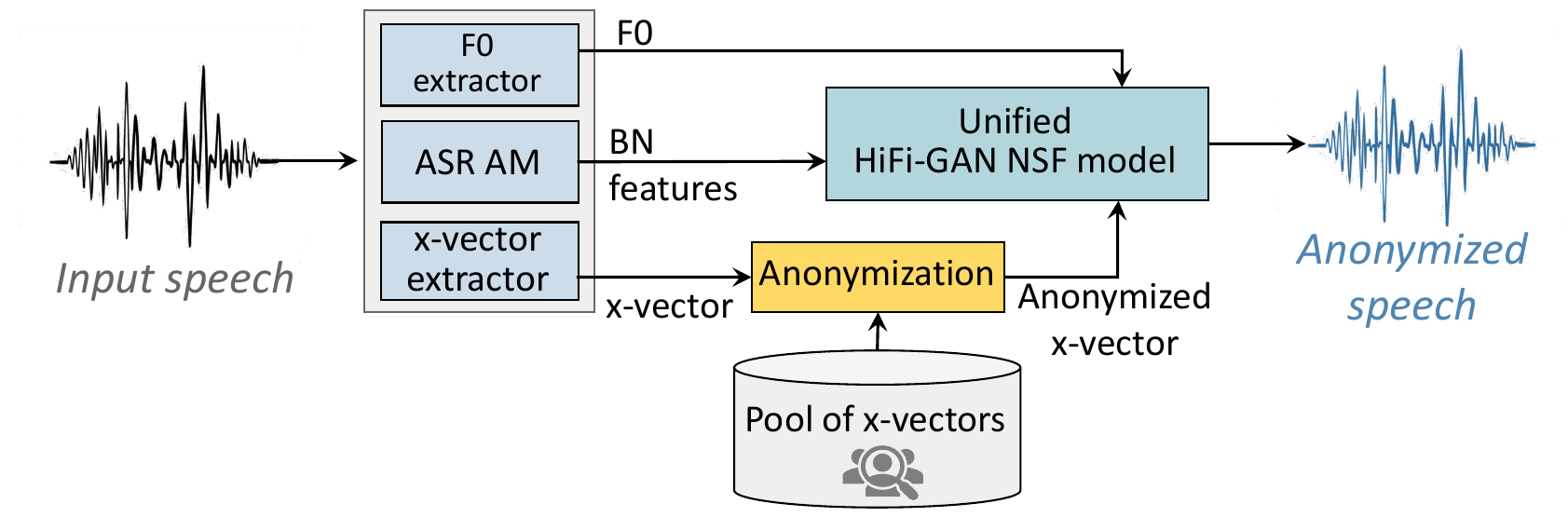}
\caption{Baseline anonymization system \textbf{B1}. 
}
\label{fig:baseline1}
\end{figure}

The \textbf{B1} baseline anonymization system \citep{srivastava2020baseline,srivastava2021}, shown in Figure \ref{fig:baseline1}, is based on the VPC 2022  \textbf{B1.b} system, but applies anonymization at the \textit{utterance level}. The system first extracts three features: 256-dimensional bottleneck (BN) features for linguistic content, fundamental frequency (F0), and 512-dimensional speaker x-vectors. BN features are obtained from a factorized time delay neural network (TDNN-F)~\citep{povey2018semi} model trained on \textit{LibriSpeech (train-clean-100, train-other-500)}, while x-vectors are from a TDNN trained on \textit{VoxCeleb-1,2}. An anonymized x-vector is then computed by averaging a random subset of 100 among the 200 farthest candidate x-vectors from a disjoint speaker pool (\textit{LibriTTS train-other-500}). Finally, the anonymized speech is synthesized with the anonymized x-vector, BN, and F0 features using a neural source-filter (NSF) model trained on \textit{LibriTTS train-clean-100} with a HiFi-GAN~\citep{kong2020hifi} architecture.

\subsection{Anonymization using the McAdams coefficient: B2}

 The \textbf{B2} baseline anonymization system~\citep{patino2020speaker} 
 uses only signal processing techniques and requires no use of training data.
 It employs the McAdams coefficient~\citep{mcadams1984spectral} to shift pole positions derived from linear predictive coding (LPC) analysis of input speech signals.
Frame-by-frame LPC analysis is used to extract coefficients and residuals. 
LPC coefficients are converted to z-plane pole positions, with 
each pole representing a spectral peak corresponding to a formant position.
The McAdams' transformation is used to modify only poles with non-zero imaginary parts by raising their phase $\phi$ (0 to $\pi$ radians) to the power of a coefficient $\alpha$, resulting in shifted phases of $\phi^\alpha$. For each utterance, $\alpha$ is sampled from $U(0.5,0.9)$, causing contraction or expansion around $\phi=1$ radian ($\approx$2.5\,kHz). 
Complex conjugate poles are shifted in opposite directions. The full set is converted back to LPC coefficients, which are combined with residuals to synthesize a frame of anonymized speech.

\subsection{Anonymization using phonetic transcriptions
and generative adversarial network:
 B3}
\label{sec:baseline_b3}
\begin{figure}[h!]
\centering\includegraphics[width=127mm]{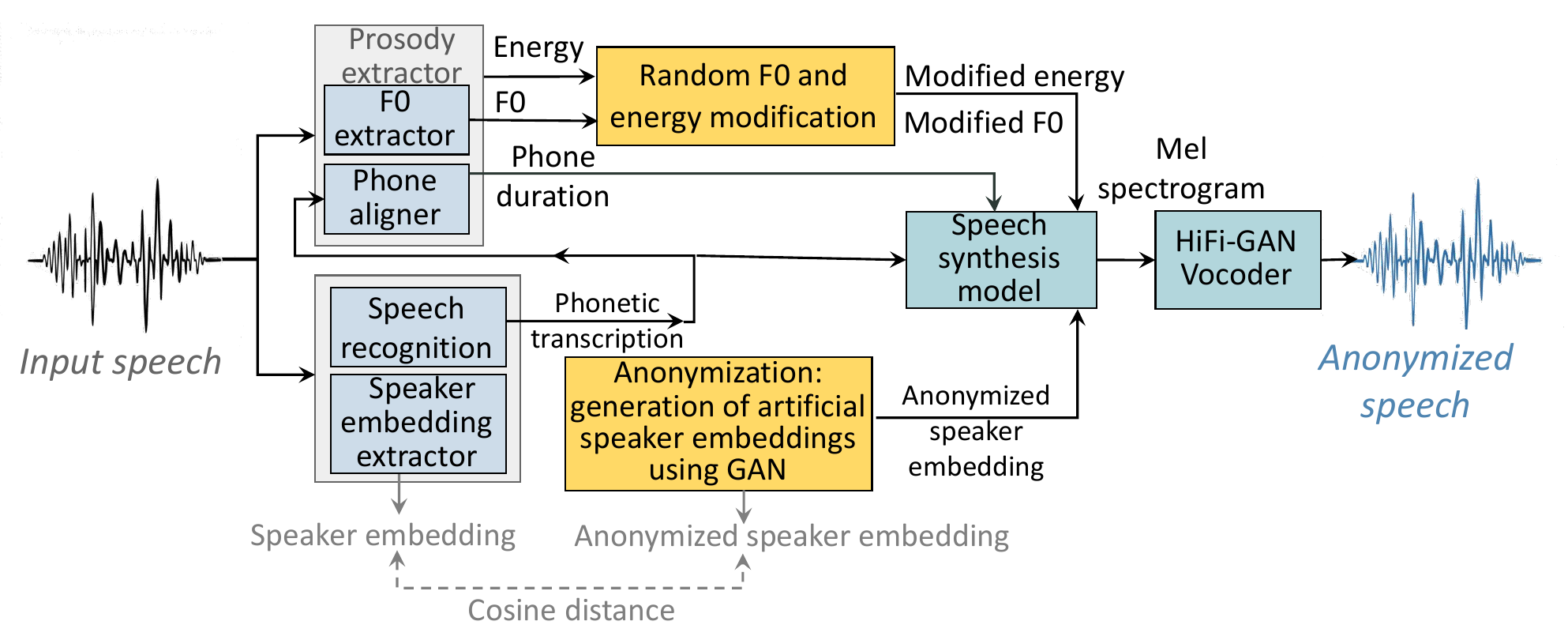}
\caption{Baseline anonymization system \textbf{B3}. 
}
\label{fig:gan_baseline}
\end{figure}

Baseline \textbf{B3}, shown in Figure~\ref{fig:gan_baseline}, is a  
speech synthesis system conditioned to retain linguistic and general prosodic information while substituting the speaker voice~\citep{meyer2023prosody,meyer2023voicepat}. 
At its core is a generative adversarial network (GAN) that generates an artificial pseudo-speaker embedding~\citep{meyer2023anonymizing}. 
\textbf{B3} performs speech anonymization through a three-step process.
First, speaker embeddings, phonetic transcriptions, the fundamental frequency (F0), energy, and phone duration are extracted from the input. 
The speaker embedding is extracted using an adapted global style tokens model~\citep{wang2018style} trained using the \textit{LibriTTS train-clean-100} dataset.
The phonetic transcription is obtained using an end-to-end ASR model with a hybrid CTC-Attention architecture, a Branchformer~\citep{peng2022branchformer} encoder and a Transformer decoder trained using the \textit{LibriTTS train-clean-100} and \textit{train-other-500} datasets.
Second, anonymization is performed by replacing the original speaker embedding with an artificial embedding generated using a Wasserstein GAN~\citep{arjovsky2017wgan} ensuring that the cosine distance between the artificial and original embeddings exceeds 0.3.  
The GAN model is trained using 
\textit{LibriTTS train-clean-100}, \textit{RAVDESS}~\citep{livingstone2018ryerson}, and \textit{ESD}~\citep{zhou2021seen} datasets. Additionally, the F0 and energy values of each phone are multiplied by random values sampled
 uniformly and independently 
from $U(0.6,1.4)$
to remove speaker-specific prosodic patterns while still retaining broader prosody dynamics. Random values are chosen for each phone individually.
Finally, the anonymized speaker embedding, 
modified F0 and energy features, original phonetic transcripts and original phone durations
are fed into a speech synthesis system based on \textit{FastSpeech2}~\citep{ren2020fastspeech} and a HiFi-GAN~\citep{kong2020hifi}, as implemented in the \textit{IMS-Toucan} Toolkit~\citep{lux2021toucan}, both trained using LibriTTS train-clean-100, to synthesize an anonymized output.

\subsection{Anonymization using neural audio codec  language modeling: B4 
}

\label{sec:baseline_b4}
\begin{figure}[h!]
\centering\includegraphics[height=4.7cm]{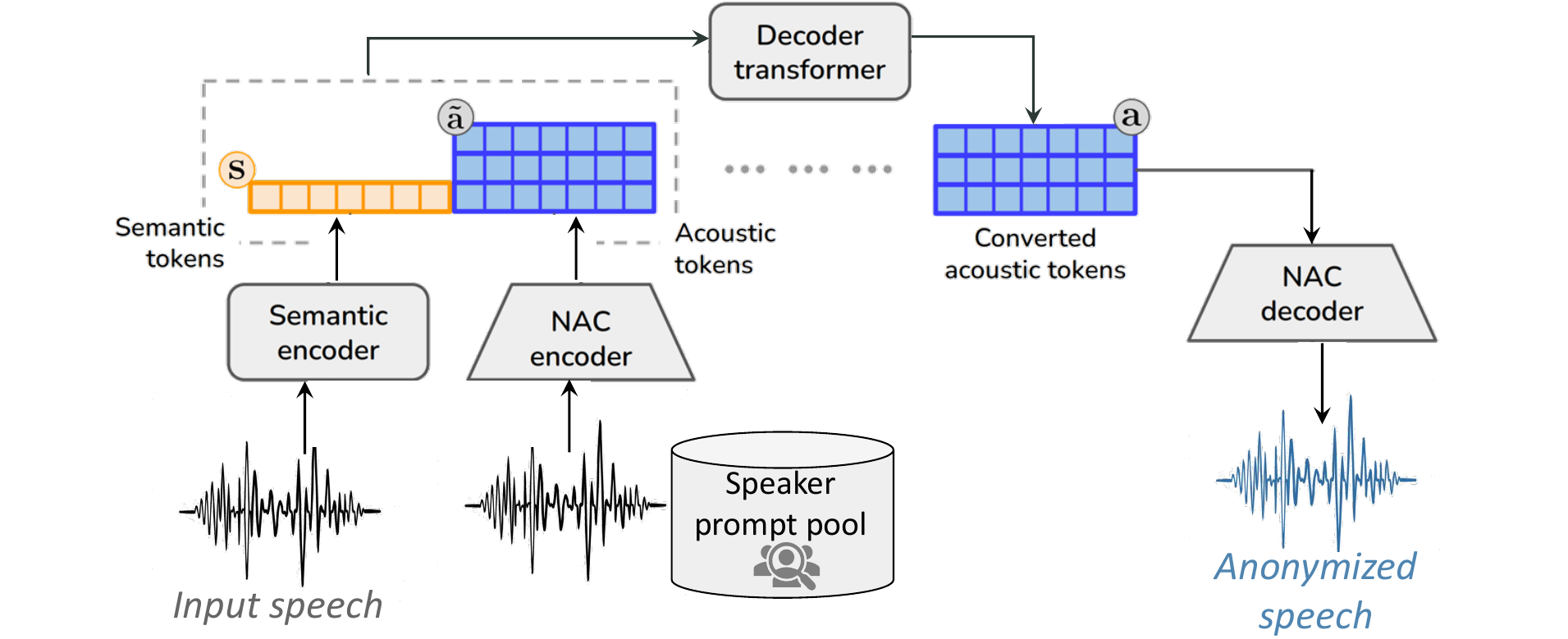}
\caption{Baseline anonymization system \textbf{B4}.}
\label{fig:B4}
\end{figure}

Baseline \textbf{B4}~\citep{panariello_speaker_2023}  illustrated in Figure~\ref{fig:B4}, is based upon the technique of neural audio codec (NAC) language modeling~\citep{audiolm, vall-e}. 
A NAC is an encoder-decoder neural network using which an input signal can be encoded into a sequence of discrete {\it acoustic tokens} $\mathbf{a} \in \{1, \dots, N_Q\}^{Q \times T_A}$ and subsequently decoded to a waveform.
$T_A$ is the number of time frames into which the input is segmented, and $Q$ is the number of tokens associated to each time frame, each drawn from one of $Q$ different token dictionaries. 
Tokens are integers ranging from $1$ to $N_Q$.
Acoustic tokens are assumed to capture the characteristics of an individual's speech. Several sets of acoustic tokens are extracted from the speech of a pool of pseudo-speakers, obtaining a {\it pool of acoustic prompts}~$\mathbf{A}$.
A \textit{semantic extractor} is used to extract a sequence of discrete \textit{semantic tokens} $\mathbf{s} \in \{1, \dots, N_S\}^{T_S}$ which encode the spoken content, where $T_S$ is the number of time frames. 
They are concatenated with a randomly chosen sequence of acoustic tokens $\mathbf{\Tilde{a}} \in \mathbf{A}$ to form  sequence $(\mathbf{s}, \mathbf{\Tilde{a}})$. 
A GPT-like, decoder-only Transformer uses this sequence as a prompt to generate acoustic tokens $\mathbf{a}$ that reflect the semantics encoded in $\mathbf{s}$ and style encoded in $\mathbf{\Tilde{a}}$. 
The NAC decoder module then converts $\mathbf{a}$ to a waveform containing the semantic content of the input, but the voice of a different pseudo-speaker.
The NAC is \textit{EnCodec}\footnote{\url{https://github.com/facebookresearch/encodec}} \citep{encodec}, which is trained with speech segments from the DNS Challenge~\citep{dubey2022icassp} and the \textit{Common Voice}~\citep{ardila2020common} dataset,  with additional non-speech segments from the \textit{AudioSet}~\citep{audioset}, \textit{FSD50K}~\citep{fsd50k}, and \textit{Jamendo}~\citep{jamendo} datasets.
The semantic extractor is composed of a HuBERT feature extractor~\citep{hubert} trained using the \textit{LibriSpeech-train-960} dataset, and a LSTM back-end that predicts token indices from feature vectors.
The decoder-only model is a publicly available checkpoint from the \textit{Bark} TTS model.\footnote{\label{footnote:bark}\url{https://github.com/suno-ai/bark}}

\subsection{Anonymization using ASR-BN with vector quantization (VQ): \textbf{B5} and B6}\label{sec:baseline_b5_b6}

\begin{figure}[h!]
\centering\includegraphics[width=90mm]{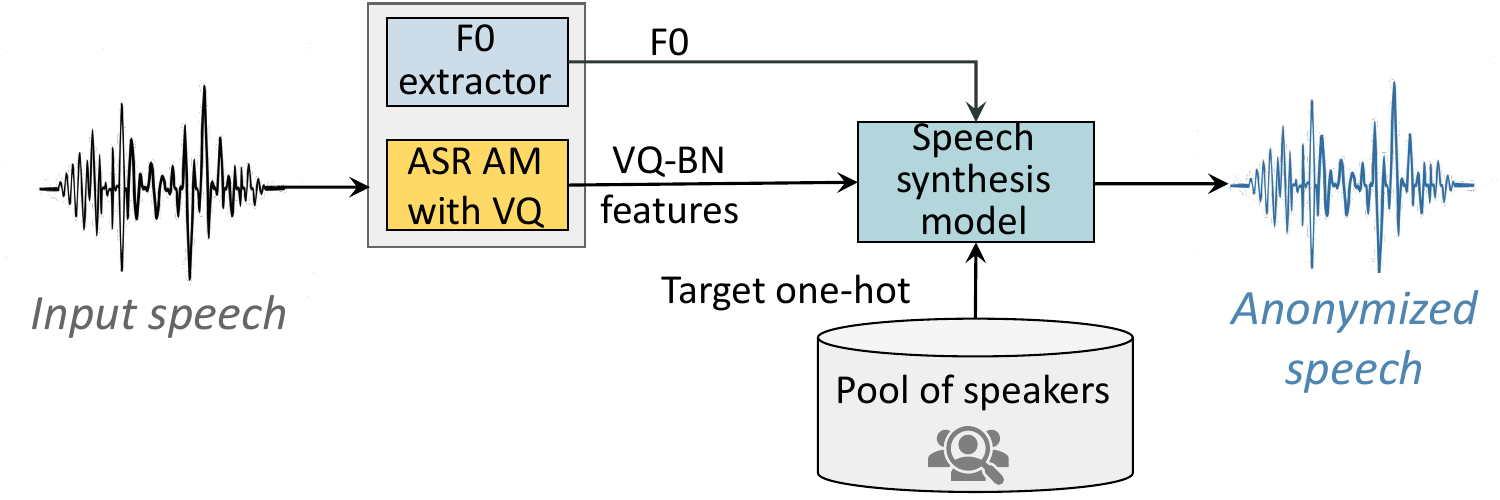}
\caption{Baseline anonymization systems \textbf{B5} and \textbf{B6}.} 
\label{fig:vq_baseline}
\end{figure}

The  pipeline for \textbf{B5} and \textbf{B6} baselines, developed by \cite{champion2023} and illustrated in  Figure~\ref{fig:vq_baseline},
shares similarities with  \textbf{B1}. 
However, it differs by incorporating vector quantization (VQ) to enhance the disentanglement of linguistic and speaker attributes. 
The pipeline extracts fundamental frequency (F0) and acoustic VQ bottleneck (VQ-BN) features from an ASR acoustic model trained to identify left-biphones. 
VQ-BN features, combined with F0 features and target speaker representation in the form of a one-hot vector corresponding to a distinct speaker in the training dataset, are then used with a HiFi-GAN network to synthesize an anonymized output. 
The use of two different ASR AMs corresponds to two different baselines:
\textbf{B5} for which the AM is a combination of a pretrained wav2vec 2.0 model with three additional TDNN-F layers; 
and
\textbf{B6} for which the AM consists  of 12 TDNN-F layers.
In the final TDNN-F layer of both models, VQ is applied following the first activation (inner bottleneck of the TDNN-F with 256 dimensions). This process approximates a continuous vector with an equivalent-dimensional vector from a finite set. The incorporation of VQ minimizes speaker information encoded within the bottleneck features.
\textbf{B5} and \textbf{B6} baselines employ multiple pre-trained models. The wav2vec 2.0 model is pre-trained using 24.1k hours of unlabeled multilingual West Germanic speech from the VoxPopuli\footnote{\url{https://dl.fbaipublicfiles.com/voxpopuli/models/wav2vec2\_large\_west\_germanic\_v2.pt}} dataset, then fine-tuned using the \textit{LibriSpeech train-clean-100} dataset. 
The 
\textbf{B6}
ASR-BN extractor operates upon Mel filterbank input feature and is trained using \textit{LibriSpeech train-other-500} and \textit{train-clean-100} datasets. Finally, the HiFi-GAN model is trained using \textit{LibriTTS-train-clean-100}, giving 247 one-hot speaker vectors.

\section{Submitted anonymization systems}
\label{sec:participants_anonym_systems}

The VPC 2024 attracted more than 100 participants from academic, industrial, and governmental organizations across 16 countries, represented by 40 teams. In total, we received 36 anonymization systems from 13 teams (listed in Table \ref{tab:teams}) who successfully submitted their results.
\begin{table}[h]
    \caption{Participant teams.}
    \centering
    \begin{tabular}{llm{8cm}}
       \toprule
        \textbf{Team name (Ref.)} & \textbf{Team ID} & \textbf{Affiliations}  \\ \midrule
        Anemone \citep{hua2024emotional} & T7 & Institute of Acoustics, Chinese Academy of Sciences; University of the Chinese Academy of Sciences, China  \\ \midrule
        JHU CLSP \citep{xinyuan2024hltcoe} & T8 & Johns Hopkins University, USA  \\ \midrule
        LongYuan \citep{tan2024system} & T9 & Nanjing University of Posts and Telecommunications; Nanjing Longyuan Information Technology Co. Ltd, China  \\ \midrule
        NPU-NTU \citep{yao2024npu}  & T10 & Audio, Speech and Language Processing Group (ASLP@NPU); School of Computer Science, Northwestern Polytechnical University; Nanyang Technological University; The Hong Kong Polytechnic University, China  \\ \midrule
        NTU-NPU \citep{kuzmin2024ntu}  & T12 & Nanyang Technological University, China; Institute for Infocomm Research, A*STAR, Singapore; Audio, Speech and Language Processing Group (ASLP@NPU), China; The Hong Kong Polytechnic University, China\\ \midrule
        ADRES \citep{leang2024exploring} & T14 & University Grenoble Alpes, CNRS, France; Institute of Digital Research and Innovation, Phnom Penh, Cambodia  \\ \midrule
        NKU HLT Lab \citep{he2024facodec}  & T17 &  College of Computer Science, Nankai University, China  \\ \midrule
        Q \citep{li2024emotion} & T18 & Qifu Technology; Fudan University, Shanghai, China \\ \midrule
        DFKI\_SLT \citep{das2024system} & T19 & Speech and Language Technology, German Research Center for Artificial Intelligence (DFKI);  Technische Universit\"at Berlin, Germany  \\ \midrule
        USTC-PolyU \citep{gu2024ustc} & T25 & NERC-SLIP, University of Science and Technology of China; The Hong Kong Polytechnic University, Hong Kong, China  \\ \midrule
        V-Beam \citep{lee2024voice,lee2024system} & T30 & Sogang University, Seoul; Ewha Womans University, Seoul, Korea  \\ \midrule
        KIT-ISL \citep{akti2024investigating,akti2024voice}  & T33 & Karlsruhe Institute of Technology (KIT), Germany; Carnegie Mellon University (CMU), USA  \\ \midrule
        Orange\_Shiva \citep{le2024orange,le2024tuning} & T38 & Orange, France  \\ \bottomrule
    \end{tabular}
    \label{tab:teams}
\end{table}
Each team submitted 1 to 6 systems that targeted different privacy categories and utility performance.
The voice anonymization solutions developed by the participants include several technical approaches. 
The summary of the systems is provided in Table \ref{tab:summary_s}. 
Noncompliance with the VPC 2024 requirements is indicated by a red asterisk (\textcolor{red}{\textbf{*}}) and further discussed in the text, where the reasons for noncompliance are \hl{highlighted in red}.

\subsection{Neural voice-conversion (VC)-based anonymization systems}

Most submitted anonymization systems are based on neural voice conversion techniques, which focus on explicit attribute disentanglement. They extract separate representations for linguistic content (e.g., bottleneck or self-supervised learning features like \textit{WavLM} or \textit{HuBERT}), speaker identity (x-vectors or ECAPA embeddings), prosody (F0), and emotion. Then, they replace speaker features with pseudo-speaker embeddings while preserving or modifying other attributes, and synthesize the speech signal via vocoders such as HiFi-GAN. 
Variants include k-nearest neighbors VC (kNN-VC), which performs frame-level nearest-neighbor replacement in \textit{WavLM} feature space, and emotion-enhanced approaches.

\subsubsection{k-nearest neighbors voice conversion (kNN-VC)}\label{subsec:kNN-VC}

Two teams, \textbf{T8} and \textbf{T19}, use k-nearest neighbor voice conversion (kNN-VC), originally proposed by~\citet{baas2023voice}.
Instead of using disentangled speaker embeddings and speaker-independent representations, kNN-VC-based methods use a \textit{WavLM} model to extract a single representation sequence from the input utterance and replace it with a `similar' representation but extracted from the utterances of another speaker selected at random from a pool. The similar representation is the average of the top $k$ nearest representations to the original. The assumption is that the similar representation encodes similar linguistic content to the input but a different speaker identity. An anonymized output is generated using a waveform generator conditioned on the resulting representations.

\paragraph{\textbf{T8-2}} 
\textit{kNN-VC system  on WavLM features} \citep{xinyuan2024hltcoe}:
A modification of the kNN-VC system which operates on \textit{WavLM} features~\citep{Chen2021WavLM}.
Source and target speaker utterances are converted into \textit{WavLM} feature space. kNN regression is then applied at the frame level, before the output is synthesized using a HiFi-GAN vocoder. 
The \textbf{T8-2} implementation uses $k=4$ and a vocoder trained using \textit{LibriSpeech train-clean-100}. 
In pursuit of improved anonymization, target speaker selection is performed at random, drawn from the \textit{VoxCeleb} corpus.

\paragraph{\textbf{T19-1}} \textit{Single-layer kNN-VC (KNNS)}~\citep{das2024system}:
Based on kNN-VC using a single-layer  \textit{WavLM} feature extraction (6th Transformer layer). The source utterance is processed through \textit{WavLM} \textit{Large} to extract frame-level features. A random target speaker is selected from the English speakers in the emotional speech database (\textit{ESD}). All utterances of the target speaker are processed by  \textit{WavLM} to create a matching set of features.
 For each source frame, the top $4$ nearest neighbors from the target speaker's feature pool are found; their average replaces the original frame. A pretrained HiFi-GAN vocoder reconstructs the audio waveform.

\paragraph{\textbf{T19-2}} \textit{ Multi-layer kNN-VC (KNND)}~\citep{das2024system}:
Extends \textbf{T19-1} by using outputs from both the 6th and 12th \textit{WavLM} layers simultaneously.
The 12th layer captures emotional cues better, enhancing emotion preservation.
Query and matching sets double in size due to this double-layer feature concatenation.
HiFi-GAN vocoder is augmented with a convolutional PreNet module, pretrained jointly for this dual-input format.

\subsubsection{Attribute disentanglement (linguistic content, speaker, pitch, emotion, etc.)}
\label{sec:disentangle}

\paragraph{\textbf{T7-1,2}} \textit{Anonymization using fusion of VC  speaker models at the model parameter level}~\citep{hua2024emotional}:
Anonymization is performed using the fusion of multiple fine-tuned, single-target speaker end-to-end VC models integrated at the parameter level. Emotion is transferred using two different emotion encoders: 
\textbf{T7-1} employs an emotion encoder based on the \textit{wav2vec 2.0} model which outputs emotion embeddings that added to the acoustic model as a priori encodings; \textbf{T7-2} leverages a global style tokens (GST)  encoder~\citep{wang2018style} composed of multi-head Attention modules.

\paragraph{\textbf{T9}} \textit{GMM-blender} \citep{tan2024system}:
A \textit{WavLM} model (layer 6 output) is used to extract
speech features. 
The \textit{GMM-Blender} method generates a rich array of anonymous templates through feature interpolation from the anonymization pool.
The \textit{GMM-Blender} operates by first randomly selecting $M$ speakers from the \textit{LibriSpeech} corpus to form a dynamic anonymization pool. 
For each speech utterance to be anonymized, four speakers are randomly selected from this pool, contributing 50 speech samples each (200 samples total). 
\textit{WavLM} embeddings are extracted from these samples and subjected to Gaussian Mixture Model (GMM) clustering, which identifies 2000 cluster centers. 
At each cluster center, the four nearest feature embeddings are selected and blended using random weights, 
creating a composite anonymized embedding template. 
A wav2vec2-based emotion feature extractor (\textit{wav2vec2-large-robust-12-ft-emotion-msp-dim}~\citep{wagner2023dawn}) is incorporated to preserve emotional characteristics during anonymization. 
Emotion features extracted from the source speech are 
fused with anonymized speech features.
This fused representation is then provided to a modified HiFi-GAN vocoder for waveform synthesis. 
The training objective combines the standard HiFi-GAN adversarial loss with an emotional similarity term that constrains the synthesized speech to maintain the emotional characteristics of the original speech.

\paragraph{\textbf{T12-5}}
\textit{\textbf{B5}-based anonymization with mean reversion pitch smoothing}
~\citep{kuzmin2024ntu}:
This system modifies the fundamental frequency (F0) in the \textbf{B5} baseline by applying a ``mean reversion'' mechanism, where the F0 estimate for each frame is interpolated with a moving average of adjacent frames. This reduces pitch variability and smooths the pitch contour to help anonymization.

\paragraph{\textbf{T12-6}}
\textit{\textbf{B5}-based anonymization with mean reversion pitch smoothing and additive Gaussian noise}
~\citep{kuzmin2024ntu}:
Similar to \textbf{T12-5}, this system additionally applies additive white Gaussian noise (AWGN) explicitly with controlled power to the mean-reverted F0 during inference. 
This further enhances privacy by increasing difficulty of speaker identification while maintaining reasonable speech utility.

\paragraph{\textbf{T14-1,2}} 
\textit{VQ variational auto-encoder (VAE) with prosody parameters}
\citep{leang2024exploring}: A custom convolutional module is used to extract content features from the input utterance which are quantized using VQ to disentangle residual speaker information from content information.
The log-normalised F0 curve of the input utterance is estimated using YAAPT~\citep{yaapt}, then fed into a bidirectional gated recurrent unit (GRU) model~\citep{gru}. 
The voice identity is encoded with an ECAPA-TDNN embedding which is anonymized 
by means of the well-known pool-based anonymization function (similar to \textbf{B1}).
A HiFi-GAN is used to synthesize the output.
\textbf{T14-1} uses standard normalized F0 and energy prosody features extracted from inputs.
\textbf{T14-2} introduces random scaling of F0 during synthesis to further obscure speaker identity. This scaling uses a random factor in range $[0.8, 1.2]$. The rest of the architecture remains consistent with \textbf{T14-1}.

\paragraph{\textbf{T18-1,2}} 
\textit{Emotion-enhanced anonymization  based on FreeVC}
\citep{li2024emotion}: 
Both systems are based on the FreeVC~\citep{li2023freevc} voice conversion framework, which uses a conditional variational autoencoder (CVAE) augmented with GAN training for one-shot voice conversion without the need for text annotations.
Both systems integrate emotional feature extraction to condition the decoder, aiming to preserve emotional content during anonymization.
\textbf{T18-1} 
uses a wav2vec 2.0 large-robust model fine-tuned on the \textit{MSP-Podcast} dataset as the emotion encoder. 
After conditioning on emotion information, the system functions in the same way as \textit{FreeVC}.
For \textbf{T18-2}, handcrafted features extracted using \textit{openSMILE}~\citep{opensmile} are processed with an LSTM neural network to create an emotion embedding.

\paragraph{\textbf{T19-3}} 
  \textit{Vector quantized mutual information-based VC (VMC)}~\citep{das2024system}:
A system based on the vector quantization and mutual information-based VC (VQMIVC) approach~\citep{wang2021vqmivc}. 
The content extractor uses VQ and a contrastive-predictive-coding-based training loss on quantized content features~\citep{wang2021vqmivc}. The training criterion acts to minimize the mutual information between the content, speaker, and pitch representations. The content and speaker extractors, both of which are convolutional DNNs, are trained jointly from scratch. The pitch extractor is based on \textit{WORLD}, while the waveform generator is an independently trained parallel WaveGAN.

\paragraph{\textbf{T25-1,2}\textcolor{red}{\textbf{*}}  \textcolor{darkgray}{and \textbf{\textit{T25-p1,p2}}}}
\textit{Emotion-aware  anonymization via content and non-content disentanglement with GST modeling}
\citep{gu2024ustc}:
Both systems share a common framework comprising:
(1) A content encoder adapted from baseline \textbf{B5} using ASR-BN features extracted via wav2vec 2.0 combined with vector quantization (VQ), generating discrete content representations;
(2) A non-content learner leveraging the GST framework to model non-content attributes like speaker identity, prosody, and emotion via a reference encoder and attention module;
(3) A HiFi-GAN vocoder that synthesizes waveform conditioned on combined content and non-content embeddings;
(4) An emotion recognition module that predicts the emotion of the input speech, used to select a reference utterance with matching emotion from a speech pool.
The anonymization pipeline works by encoding content and emotion from the original speech, selecting a reference utterance with the same emotion for pseudo-speaker embedding extraction, and synthesizing anonymized speech with the combination of original content and pseudo non-content information.
The two submitted systems mainly differ in composition of the emotion-labeled corpora for non-content learner:
In \textbf{T25-1}, \textit{LibriTTS train-clean-100} datasets supplemented with English utterances from the \textit{ESD}; in  \textbf{T25-2}, only the \textit{ESD} part was used.
\hl{SER (fine-tuned on \textit{IEMOCAP}) is used at test time to select utterances with matching emotions.}
However, post-evaluation analysis by this team showed that the rule violation was not critical. Specifically, the authors evaluated these anonymization systems using an alternative SER model: the MMER model (a multimodal multi-task learning approach for speech emotion recognition~\citep{ghosh23b_interspeech}).
The MMER model was pre-trained on the \textit{MSP-Podcast} corpus and fine-tuned on the \textit{ESD} dataset.
The corresponding systems \textbf{T25-p1} and \textbf{T25-p2}, which replace the \textit{IEMOCAP} SER with MMER, comply with challenge rules.
\textbf{T25-p1} (like \textbf{T25-1}) uses \textit{ESD} as the reference pool, while \textbf{T25-p2} uses both \textit{ESD} and \textit{LibriTTS train-clean-100}.
Results for \textbf{T25-p1} and \textbf{T25-p2} show that MMER improves UAR without significantly degrading other metrics (see Table~\ref{tab:results}). 
%%%%%%%%

\paragraph{\textbf{T33-1,2}} 
\textit{FreeVC-based VC systems with HuBERT and ASR  models}
\citep{akti2024investigating,akti2024voice}:
A zero-shot VC-system based on \textit{FreeVC}~\citep{li2023freevc}:
Content features are extracted from the final layer of a transformer-based ASR system and combined  with a target speaker embedding and F0 estimates before waveform generation.
The target speaker embedding 
is selected at random from the VCTK training set.  \textbf{T33-1} is similar to \textbf{T33-2} but uses a pre-trained \textit{HuBERT Base} model instead of the transformer-based ASR. The outputs from the 9th transformer block of the HuBERT are used as the content features.

\paragraph{\textbf{T38-1,2,3,4,5}}
\textit{DISSC-based systems}
\citep{le2024orange,le2024tuning}:
All based on the use of DIscrete units for Speaking
Style Conversion (DISSC)~\citep{maimon-adi-2023-speaking}, an approach to VC which uses disentangled SSL representations.  
\textbf{T38-1} is the original DISSC system. 
Discrete HuBERT representations, quantized using k-means to encode linguistic content, are input to predictors that estimate the duration of each discrete unit and F0.
Content features and a target speaker embedding randomly selected from the VCTK training set are used by HiFi-GAN to synthesize the output. 
\textbf{T38-2} uses durations and F0 extracted from the input waveform. 
Extending the same system and to retain expressiveness, 
\textbf{T38-3} uses a HiFi-GAN fine-tuned using the \textit{MSP-Podcast} training set.
\textbf{T38-4} adds random perturbations to F0 estimates.
\textbf{T38-5}, a variant of \textbf{T38-1}, uses \textit{MSP-Podcast} data to fine-tune duration and F0 prediction modules. 
Finally, \textbf{T38-6} extends \textbf{T38-5} by concatenating speaker embeddings with emotion embeddings extracted from the source waveform using a pre-trained SSL model.

\subsection{Anonymization systems based on neural audio codecs}
Neural codec-based voice anonymization systems represent a modern approach leveraging the powerful representation capabilities of neural audio codecs (NACs). These codecs transform speech into quantized codes that effectively bottleneck speaker-related information while capturing linguistic content and prosodic features. By encoding speech into discrete bottleneck representations, these systems inherently disentangle speaker identity, linguistic content, and prosody. Anonymization is achieved by manipulating or replacing the speaker identity component at the bottleneck level while preserving content and prosodic cues to maintain speech naturalness and intelligibility. Compared to most traditional VC methods that rely on explicit feature extraction and modification, neural codec systems operate within a unified codec framework that naturally separates and recombines speech components.

\paragraph{\textbf{T10-1,2}\textcolor{red}{\textbf{*}}} 
\textit{Disentangled neural codec  anonymization  with serial disentanglement and multi-level distillation}
\citep{yao2024npu}:
A pair of neural codec systems that use disentangled representations of speaker identity, linguistic content, and fundamental frequency. Emotion cues are preserved through frame-level emotion distillation, while linguistic content and speaker information are disentangled via semantic teacher and self-supervised speaker distillation. Anonymization is performed using a weighted sum of an average voice identity derived from a speaker pool and a random identity selected from a Gaussian distribution.
The parameter $\alpha$ in the anonymization system is a weight that controls the mixture between an averaged speaker identity vector and a randomly sampled speaker identity vector during the anonymization process of speaker identity:
$\mathbf{s}_{\text{anon}} = \alpha \cdot \bar{\mathbf{s}} + (1 - \alpha) \cdot \hat{\mathbf{s}}$,
where:
    $\mathbf{s}_{\text{anon}}$ is the anonymized speaker identity vector,
     $\bar{\mathbf{s}}$ is the averaged speaker identity vector computed from a selected set of candidate speaker identities,
     $\hat{\mathbf{s}}$ is a randomly generated speaker identity vector,
     $\alpha \in [0,1]$ is the mixing weight controlling the degree of influence of the averaged and random vectors.
A higher $\alpha$ puts more weight on the averaged speaker identity, typically resulting in less anonymity but better utility preservation, while a lower $\alpha$ increases randomness, enhancing anonymity but potentially decreasing utility.
The parameter $\alpha$ allows adjustment of the privacy category. Both systems differ in their weights and belong to different privacy categories: \textbf{T10-1} with $\alpha = 0.9$ belongs to the third privacy category (EER $\geq$ 30\%), and \textbf{T10-2} with $\alpha = 0.8$ belongs to the last one (EER $\geq$ 40\%).
\hl{These systems fail to fully comply with challenge requirements for anonymizing training data used in ASV model training. Specifically, training data should be anonymized at the \textit{utterance level} using the same randomization strategy as applied to test data. Deviations from this protocol may result in a weaker attack model and consequently overestimate the actual level of privacy protection achieved by these systems.}

\paragraph{\textbf{T12-1}} 
\textit{FACodec-based  anonymization with pool averaging, cross-gender conversion, and Gaussian noise}
\citep{kuzmin2024ntu}: A factorized neural speech codec (FACodec)~\citep{ju2024naturalspeech} from NaturalSpeech3 is used to disentangle prosody, content, and other acoustic detail. The codec is conditioned on a speaker embedding generated by a timbre extractor.
Anonymization is performed by replacing the original speaker embedding extracted by the codec with an anonymized one. The anonymization strategies experimented with include: pool-based averaging, cross-gender conversion, and the addition of white Gaussian noise (AWGN) to the speaker embedding with varying noise power levels.

\paragraph{\textbf{T17}\textcolor{red}{\textbf{*}}}
\textit{FACodec-based  anonymization with enhanced prosody transformation}
\citep{he2024facodec}: A FACodec is modified with an additional prosody anonymization module which operates upon a Mel spectrogram and a target speaker embedding. 
Speaker-specific features contained within latent prosodic features generated using a prosody decoder
is suppressed using a speaker embedding extractor paired with a gradient reversal layer. A prosody decoder generates a new Mel spectrogram which contains the prosodic information of the target speaker.
The timbre embedding extracted from the original utterance is also replaced with that of the target speaker.
\hl{Computing a center of speaker embeddings across the entire dataset before anonymization, where embeddings are extracted by the FACodec timbre extractor, may violate the VPC 2024 requirement that trial utterances be processed independently of each other.}

\subsection{Cascaded ASR+TTS anonymization systems}

This section covers anonymization techniques that employ a sequential pipeline of ASR followed by text-to-speech (TTS) synthesis. These cascaded systems transcribe speech into word or phonetic sequences as an intermediate representation, which is then resynthesized to produce anonymized voice outputs.

\subsubsection{Conventional cascaded ASR-TTS}

These systems follow the traditional approach of passing text transcripts from ASR directly to TTS without explicitly preserving prosodic or paralinguistic attributes. 

\paragraph{\textbf{T8-1}}  \textit{Cascaded ASR (Whisper) and TTS (VITS)} \citep{xinyuan2024hltcoe}:
  Text transcripts are derived from the input
using the \textit{medium English Whisper} model. A VITS~\citep{kim2021conditional} TTS model trained using the \textit{LibriTTS} dataset is then used to generate anonymized outputs.

\subsubsection{Cascaded ASR-TTS systems with paralinguistic preservation}

These advanced cascaded systems integrate mechanisms to preserve paralinguistic features such as emotion, intonation, and speaking style alongside speech content.

\paragraph{\textbf{T12-2,3}} 
\textit{{\textbf{B3}-based system with integrated emotion embeddings and cross-gender
averaging and selecting anonymization}}
\citep{kuzmin2024ntu}: Both systems are based on the \textbf{B3} baseline system and utilize the \textit{FastSpeech2} model with HiFi-GAN vocoder to synthesize speech. They integrate emotion embeddings extracted with a fine-tuned \textit{wav2vec 2.0} model to enhance emotion 
preservation. Speaker anonymization in both cases involves cross-gender related strategies.
The  differences lie in their speaker anonymization mechanisms and embedding usage,  and prosody modification module. 
System \textbf{T12-2} anonymizes the speaker by averaging 100 speaker embeddings randomly selected from a pool of 200 farthest embeddings from the source speaker, combined with cross-gender selection. System \textbf{T12-3} employs random speaker selection from a cross-gender pool on a per-utterance basis rather than embedding averaging. In addition, \textbf{T12-3} modifies prosody through pitch and energy features normalized and adjusted by multiplicative random factors in the range $[0.7,1.3]$.

\paragraph{\textbf{T12-4}}
\textit{\textbf{B3}-based anonymization with cross-gender selection-based anonymization}
~\citep{kuzmin2024ntu}: This system is based on the \textbf{B3} baseline but does not modify prosody. Instead of using GAN-generated pseudo-speaker embeddings, it applies cross-gender selection-based anonymization for speaker embedding anonymization.

\paragraph{\textbf{T30-1}}
\textit{\textbf{B3}-based anonymization with embedding pool perturbation and  k-means clustering}
\citep{lee2024voice,lee2024system}:
The GAN-based speaker embedding generation process of \textbf{B3} is replaced with a pool-based anonymization algorithm. 
The speaker embedding from the pool with the largest cosine distance to the input speaker embedding is selected. A random selection of embedding dimensions is then replaced with the corresponding dimensions from a speaker embedding selected at random from the 500 pool cluster centroids.

\paragraph{\textbf{T30-2}}
\textit{Cascaded ASR+TTS anonymization with emotionally enriched feature integration}
\citep{lee2024voice,lee2024system}:
The concatenation of Whisper ASR and variational inference with adversarial learning for end-to-end text-to-speech (VITS)~\citep{kim2021conditional} TTS systems. F0 estimates extracted from the input waveform are transformed into latent features which are then combined with a set of emotion prototype vectors using cross attention. Combined features are input to a Feature-wise Linear Modulation (FiLM) layer, the output of which is multiplied with Whisper ASR text embeddings. The resulting features are used with VITS for waveform generation.

\subsection{Hybrid anonymization systems}

Hybrid systems combine multiple anonymization techniques to balance privacy and utility. Different approaches exhibit distinct trade-offs: cascaded ASR+TTS achieves strong anonymization but degrades emotion preservation, while VC preserves paralinguistic content at the cost of weaker privacy. Hybrid systems address this by enabling dynamic method selection, such as the randomized \textit{admixture} approach proposed in challenge submissions, allowing fine-grained control over the privacy-utility trade-off.

\paragraph{\textbf{T8-3,4,5}} \textit{Admixture}  \citep{xinyuan2024hltcoe}:
Anonymization is carried out according to one of two methods selected at random. 
The first is the cascaded ASR-TTS system \textbf{T8-1} whereas the second is the 
kNN-VC system \textbf{T8-2}. 
This hybrid method provides a balance between anonymization strength and the preservation of utility, with the \textbf{T8-1} system better suppressing voice information at the cost of reduced  utility and the \textbf{T8-2} system better preserving linguistic and emotional content at the cost of reduced anonymization.

\begin{longtable}[htbp!]{lm{3cm}m{4.0cm}m{7.0cm}}

\caption{Summary of the submitted systems. Colors in the first column highlight the privacy category of the corresponding systems: 
\fcolorbox{black}{Orange!20}{\rule{0pt}{4pt}\rule{4pt}{0pt}} EER$\geq10\%$,
\fcolorbox{black}{Emerald!10}{\rule{0pt}{4pt}\rule{4pt}{0pt}} EER$\geq20\%$,
\fcolorbox{black}{Emerald!30}{\rule{0pt}{4pt}\rule{4pt}{0pt}} EER$\geq30\%$,
\fcolorbox{black}{Emerald!70}{\rule{0pt}{4pt}\rule{4pt}{0pt}} EER$\geq40\%$.
 Noncompliance with VPC 2024 requirements is marked by\textcolor{red}{\textbf{*}}.
}
\small
\label{tab:summary_s} \\
        \toprule
        \textbf{ID} & \textbf{ID (paper)} & \textbf{Training resources} & \textbf{Description} \\
        \midrule
        \endhead
        \cellcolor{Emerald!30}{
            \textbf{T7-1}}  &  ppg-w2vF0-fusion             
            & LibriTTS train-clean-360,
            
            LibriTTS  train-other-500, 
            
            LJSpeech, 
            ESD,
            
            wav2vec 2.0 FT on MSP-Podcast 
        
            & anonymization: fusion at the parameter level of pseudo-speakers (several pretrained end-to-end VC models fine-tuned with single speaker data); 
            
            emotion encoder: based on wav2vec 2.0; 
            F0: Yaapt; 
            linguistic content: PPG-features from a hybrid Transformer-CTC ASR  \\ \midrule
             \cellcolor{Orange!30}{\textbf{T7-2}}   & ppg-gstF0-fusion             
             &  LibriTTS train-clean-360, 
             
             LibriTTS train-other-500, 
             
             LJSpeech, 
             ESD 
             & similar to \textbf{T7-1} except for emotion encoder: GST  \\ \midrule
             \cellcolor{Emerald!70}{\textbf{T8-1}}   & Whisper-VITS TTS           
             & LibriTTS,
             
             medium English Whisper   
             & cascaded ASR (Whisper) and TTS (VITS) \\ \midrule
             \cellcolor{Gray!5}{\textbf{T8-2}}   & kNN-VC                       &
             LibriSpeech train-clean-100, 
             VoxCeleb;
             WavLM &
             kNN-VC system  on WavLM features \\ \midrule
             \cellcolor{Emerald!10}{\textbf{T8-3}}   & Admixture ($p=0.2$)       &     &  \multirow{3}{*}{\parbox{6.9cm}{random selection of one of two methods for each utterance (with probability $p$ for the second method): \textbf{T8-1} (cascaded ASR-TTS)   and \textbf{T8-2} (kNN-VC)  }}\\ \cmidrule{1-2}
             \cellcolor{Emerald!30}{\textbf{T8-4}}   & \parbox{3.3cm}{Admixture ($p=0.325$)}          & as in \{\textbf{T8-1} $\cup$  \textbf{T8-2}\} &  \\
             \cmidrule{1-2}
             \cellcolor{Emerald!70}{\textbf{T8-5}}   & Admixture ($p=0.4$)            &  &  \\ \midrule
             \cellcolor{Emerald!30}{\textbf{T9}}     & GMM-Blender                 &   ESD,
             LibriSpeech train-clean-360, 
             train-other-500, 
             WavLM,  
             wav2vec2-large-robust-12-ft-emotion-msp-dim
             &   anonymization: \textit{GMM-Blender} based on GMM speaker clustering to generate anonymous templates; 
             emotion encoder: wav2vec2-large-robust-12-ft-emotion-msp-dim  \\ \midrule
             \cellcolor{Emerald!30}{\textbf{{T10-1}\textcolor{red}{*}}}  &  C3                          &  \multirow{2}{*}{\parbox{4.5cm}{LibriSpeech, \\LibriTTS; \\WavLM}} 
             &  {\parbox{6.9cm}{neural audio codec, with a specific disentanglement strategy;
             anonymization: weighted sum of an averaged speaker identity (from speakers in a speaker pool) and a randomly generated)}}\\  \cmidrule{1-2} \cmidrule{4-4}
            \cellcolor{Emerald!70}{ 
             \textbf{{T10-2}}\textcolor{red}{*}}  & C4                           &    & similar to \textbf{T10-1}, but with smaller weight for an averaged speaker identity  \\ \midrule
             \cellcolor{Gray!5}{\textbf{T12-1}}  & 1a                           &  \parbox{4.0cm}{LibriTTS train-clean-100, \\ LibriLight train, \\                            NaturalSpeech3 FACodec} &  relies on FACodec,

             anonymization:  pool-based, cross-gender selection,  addition of white Gaussian noise. \\ 
             
             \midrule 
             \cellcolor{Orange!30}{\textbf{T12-2}}  & 1b                           &  \parbox{4.0cm}{LibriTTS train-clean-100, \\ train-other-500, \\ MSP-Podcast;  wav2vec 2.0
             }  & modifies \textbf{B3};
             emotion encoder: wav2vec-based;

             anonymization: pool-based, selection+averaging, 
             cosine distance, cross-gender)
             \\  \midrule
             \cellcolor{Emerald!10}{\textbf{T12-3}}  & 2a                           & {\parbox{4.0cm}{LibriTTS train-clean-100, \\ train-other-500, \\ MSP-Podcast; \\ wav2vec 2.0

             }}  & modifies \textbf{B3};
             emotion encoder:  FT wav2vec 2.0 
             prosody:
             value-wise F0 and energy multiplication by random values;

             anonymization: random selection, cross-gender \\  \midrule
             \cellcolor{Emerald!10}{\textbf{T12-4}}  & 2b                           &  {\parbox{4.0cm}{LibriTTS train-clean-100, \\ train-other-500
             }} & modifies \textbf{B3}; 
             anonymization: random selection, cross-gender   \\ \midrule
             \cellcolor{Emerald!70}{\textbf{T12-5}}  & 3                            &  \multirow{2}{*}{\parbox{4.0cm}{  VoxPopuli, \\ Librispeech train-clean-100, \\ LibriTTS train-clean-100}}
             & modifies \textbf{B5};
             prosody: replaces the pitch curve with a weighted average between the curve itself and its moving average. \\  \cmidrule{1-2} \cmidrule{4-4}
             \cellcolor{Emerald!30}{\textbf{T12-6}}  & 4         & 
             & similar to  \textbf{T12-5},  adds white Gaussian noise to the resulting pitch curve. 
             \\  \midrule
             \cellcolor{Gray!5}{\textbf{T14-1}}  & v1                           &  \multirow{2}{*}{\parbox{4.5cm}{LibriSpeech, \\ CREMA-D}}  & 
             VQ VAE;
             prosody: normalized F0 and energy  from inputs;
             anonymization: selection+averaging as in \textbf{B1}  
             \\   \cmidrule{1-2} \cmidrule{4-4}
              \cellcolor{Gray!5}{\textbf{T14-2}}  & v2                           &   & similar to \textbf{T14-1} with  F0  scaling by random factors in range $[0.8, 1.2]$. \\ \midrule
             \cellcolor{Emerald!10}{\textbf{{\textbf{T17}}\textcolor{red}{*}}}    & FACodec+PANO                      & LibriLight,
             
             LibriSpeech train-other-500; 
             
             VITS (for prosody encoder); 
             
             NaturalSpeech3 codec & adds a prosody anonymization module to FACodec.
             \\ \midrule
             \cellcolor{Gray!5}{{\textbf{T18-1}}}  & EESA(Wv)                     &  MSP-Podcast,
             RAVDESS,
             VCTK;
             FreeVC, 
             WavLM, 
             wav2vec~2.0, 
             wav2vec2-large-robust12-ft-emotion-mspdim & FreeVC integrated with emotion embeddings generated by a fine-tuned version of wav2vec 2.0. \\ \midrule
             \cellcolor{Gray!5}{\textbf{T18-2}}  & EESA(OS)                      & MSP-Podcast,
             RAVDESS; 
             VCTK,
             FreeVC,
             WavLM,
             
             wav2vec 2.0   & 
             FreeVC enhanced with emotion embeddings extracted via an LSTM network operating on openSMILE features
             \\ \midrule
             \cellcolor{Gray!5}{\textbf{T19-1}}  & KNNS  & \multirow{3}{*}{\parbox{4.1cm}{LibriSpeech train-clean-360, 
             \\ ESD, \\ RAVDESS, \\ CREMA-D}} &  kNN-VC system using WavLM features (from 6th Transformer block) \\ \cmidrule{1-2} \cmidrule{4-4}
             \cellcolor{Gray!5}{\textbf{T19-2}}  & KNND  & & kNN-VC system using WavLM features (concatenated from the 6th \& 12th Transformer blocks) \\ \cmidrule{1-2} \cmidrule{4-4}
             \cellcolor{Orange!30}{\textbf{T19-3}}  & VMC   &  &  anonymization: randomly selected speaker vector from ESD speakers; 
             F0: PyWORLD; 
             
             linguistic content: CNN + vector quantization. \\ \midrule
        \cellcolor{Emerald!70}{
             \textbf{{T25-1}}\textcolor{red}{*}}  & large: ESD+LibriTTS          &  \multirow{2}{*}{\parbox{4.0cm}{ESD, \\ LibriTTS train-clean-100, \\
             \hl{IEMOCAP (test time)};  \\ wav2vec 2.0 \\
             content encoder from \textbf{B5} }}& disentanglement of content (VQ-BN as in
\textbf{B5}) and style  features; emotion transfer from target speaker utterances;
non-content learner: trained on LibriTTS train-clean-100 datasets supplemented with English utterances from
the ESD;
\\ \cmidrule{1-2} \cmidrule{4-4}
             \cellcolor{Emerald!30}{\textbf{{T25-2}}}\textcolor{red}{*}  & small:
             ESD only              &   & similar to \textbf{T25-1}, but the non-content learner is trained on
the ESD subset only \\ \midrule
             \cellcolor{Emerald!10}{\textbf{T30-1}}  & V-Beam\_sttts-tj-method3   &  LibriTTS, 
             RAVDESS, 
             ESD & 
             modifies \textbf{B3};
             anonymization: 
             selection-based from a pool created by k-means clustering 
             \\ \midrule
             \cellcolor{Emerald!70}{\textbf{T30-2}}  & V-Beam\_method2            &  LJspeech; 
             Whisper,
             VITS  & cascaded ASR (Whisper) + TTS (VITS using LJspeech)
             with  integration of emotion class prototype features with  F0
              \\  \midrule
               \cellcolor{Orange!30}{\textbf{T33-1}}  & freevc-hubert-base-ss        & \parbox{4.5cm}{VCTK; \\ HuBERT Base L 
               
               (km500 quantizer)
               }&  anonymization: randomly selected from VCTK; 
               
               F0: not explicitly extracted or converted;  
               
               linguistic content: features extracted from the 9th block of HuBERT \\  \midrule
             \cellcolor{Emerald!30}{\textbf{T33-2}}  & freevc-asr-bottleneck-f0-ss  &  \parbox{4.0cm}{  VCTK, \\LibriSpeech} & similar to \textbf{T33-1}, but  content features are extracted from the final layer of a CTC-Transformer-based ASR
             \\ \midrule
             \cellcolor{Emerald!10}{\textbf{T38-1}}  & M0 - original DISSC          & DISSC model trained on VCTK  & DISSC using speaker embeddings randomly selected from the VCTK speakers  \\ \midrule
             \cellcolor{Orange!30}{\textbf{T38-2}}  & M1 - prosody preservation    &  as in \textbf{T38-1} & \textbf{T38-1} but using duration and F0 of input waveform \\ \midrule
             \cellcolor{Orange!30}{\textbf{T38-3}}  & M2 - dataset for expressiveness &  \multirow{3}{*}{\parbox{4.0cm}{as in \textbf{T38-1} + MSP-Podcast}} & \textbf{T38-1} + HiFi-GAN trained on MSP-Podcast \\ \cmidrule{1-2} \cmidrule{4-4}
             \cellcolor{Orange!30}{\textbf{T38-4}}  & M3 - add  randomeness to prosody &  &  \textbf{T38-3} + random noise in F0 \\ \cmidrule{1-2} \cmidrule{4-4}
             \cellcolor{Emerald!30}{\textbf{T38-5}}  & M4 - prosody prediction with MSP-Podcast &  & \textbf{T38-1} + prosody predictors trained on MSP-Podcast \\ \midrule
             \cellcolor{Emerald!30}{\textbf{T38-6}}  & M5 - emotion embeddings &  
             MSP-Podcast; 
             DISSC model trained on VCTK,
             
             wav2vec2-large-robust-12-ftemotion-msp-dim  &  \textbf{T38-5} + emotion embeddings 
             \\
        \bottomrule
\end{longtable}
\normalsize

\subsection{Trends}

\subsubsection{Anonymization}

\paragraph{Cascade-based approaches:}
Systems like \textbf{T8-1} that cascade ASR+TTS achieve near-perfect anonymization with an EER of approximately 48\%. 
However, while linguistic content is largely preserved, there are
%achieving such strong privacy requirements 
significant compromises on emotion preservation. 
To achieve a better balance,
\textit{admixture} strategies were employed whereby the specific anonymization method applied to each utterance is selected at random, allowing a flexible privacy-utility trade-off. 
The benefit of such an approach is seen in 
results for hybrid systems \textbf{T8-3}, \textbf{T8-4}, \textbf{T8-5} that apply cascaded ASR-TTS and kNN-VC methods to each utterance at random.

\paragraph{Speaker embedding anonymization methods:}

\begin{itemize}
    
\item
\textit{Weighted averaging and random mixing:} 
 \textbf{T10-1,2} systems combine an averaged speaker identity derived from a speaker pool with a randomly generated speaker identity sampled from a Gaussian distribution. A mixing weight parameter controls the balance between averaged and random components. Higher weights better preserve utility by favoring averaged identities, while lower weights increase randomness for enhanced privacy.

\item
\textit{Pool-based selection with optional averaging and cross-gender conversion:} 
Multiple systems employ pool-based speaker embedding selection strategies with varying degrees of averaging and cross-gender constraints:

\begin{itemize}
\item
\textit{Averaging-based strategies}:
\textbf{T12-1} implements the FACodec-based system with pool-based speaker embedding averaging combined with cross-gender speaker selection and the injection of additive white Gaussian noise to speaker embeddings with varying noise power levels. \textbf{T14-1,2} employ selection and averaging as in baseline \textbf{B1}.
Also similar to \textbf{B1}, \textbf{T12-2} averages 100 speaker embeddings selected at random from a pool of 200 farthest embeddings based on cosine distance, combined with cross-gender selection.

\item
\textit{Single speaker selection strategies:}
\textbf{T12-3} employs random speaker selection from a cross-gender pool on a per-utterance basis.
\textbf{T30-1} uses selection-based anonymization from a pool created by k-means clustering of speaker embeddings. 
\textbf{T33-1,2} selects single pseudo-speaker embeddings at random from the VCTK speaker corpus. 
\textbf{T38-1,2,3,4,5,6} implements DISSC using speaker embeddings selected at random from VCTK speakers.

\end{itemize}

\item 
\textit{Emotion-matched pseudo-speaker selection:}
\textbf{T25-1,2} systems employ an emotion-aware anonymization strategy with which pseudo-speakers are selected from emotion-labeled pools. An emotion recognition module is used to estimate the emotion of the input speech so that a reference utterance with matching emotion is selected from the pool. The strategy helps to better maintain emotional consistency while still replacing the voice identity.

\end{itemize}

\paragraph{F0 manipulation for prosody alteration:}

\textbf{T12-3} modifies prosody through value-wise F0 and energy scaling by random multiplication factors. \textbf{T12-5,6} achieve prosodic anonymization through mean-reversion pitch smoothing, where F0 estimates are interpolated with the moving average of adjacent frames. \textbf{T12-6} additionally applies additive white Gaussian noise  to the smoothed F0 estimates. \textbf{T14-2} introduces random F0 scaling  by factors in the range 0.8--1.2 
during synthesis to obscure pitch-related speaker cues. \textbf{T17} integrates  an additional prosody anonymization module.

\subsubsection{Emotion preservation}

Diverse strategies were explored in VPC 2024 submissions to improve upon the preservation of emotion cues during voice anonymization.

\begin{itemize}
    \item \textit{Pretrained SSL-based emotion encoders:}
\textbf{T7-1} and \textbf{T18-1} employ \textit{wav2vec 2.0} models fine-tuned on emotional corpora (MSP-Podcast, ESD) to extract emotion embeddings that condition the decoder during synthesis.
\textbf{T19-2} extends single-layer \textit{WavLM} features (6th Transformer block) to dual-layer concatenation of the 6th and 12th blocks, with the 12th layer capturing emotional cues more effectively.

\item \textit{Handcrafted feature-based emotion encoding}:
\textbf{T18-2}
uses \textit{openSMILE} to extract acoustic features 
which are processed with an LSTM network to generate emotion embeddings.

\item \textit{Global Style Token (GST) approaches}:
\textbf{T7-2} and \textbf{T25-1,2} leverage GST frameworks composed of multi-head Attention modules to model emotion and style features as learnable style embeddings extracted through reference encoders.

\item \textit{Emotion distillation}:
\textbf{T10}'s neural codec systems use frame-level emotion distillation as part of a multi-stage disentanglement strategy, explicitly preserving emotional information alongside linguistic content and speaker identity.

\item \textit{Emotion recognition-guided reference selection:}
\textbf{T25-1,2} use a SER model trained on \textit{IEMOCAP} to estimate the emotion in the input speech. Reference utterances with matching emotion labels are then selected from emotion-annotated pools (\textit{ESD} or \textit{ESD+LibriTTS}), to better preserve emotional expressiveness.

\end{itemize}

\subsubsection{Linguistic content preservation}

Content preservation strategies vary based on the anonymization approach.

\begin{itemize}

\item
\textit{Vector quantization (VQ) bottleneck features:}
Multiple systems (\textbf{T14-1,2}, \textbf{T19-3}, \textbf{T12-5,6}, and \textbf{T25-1,2}) employ vector quantization-based disentanglement to separate speaker identity from linguistic content. 
\textbf{T14-1,2} use a  convolutional module to extract content features, which are subsequently quantized via VQ to remove residual speaker information while preserving content representation. 
Similar to baselines \textbf{B5} and \textbf{B6}, \textbf{T12-5,6} and \textbf{T25-1,2} apply VQ to ASR-BN features.  Continuous bottleneck representations are discretized into tokens to reduce speaker leakage.
The principle underlying these approaches is that quantization into discrete tokens acts as a bottleneck that attenuates  speaker-specific variations while retaining the discrete linguistic content necessary for speech recognition and synthesis.

\item
\textit{Pretrained SSL representations}:
Several systems leverage pretrained self-supervised speech models as content encoders, exploiting their robust capture of linguistic information across diverse acoustic conditions. \textbf{T33-1} extracts content features from the \textit{HuBERT} \textit{Base} model; \textbf{T8-2} and \textbf{T19-1,2} employ \textit{WavLM} features.
\textbf{T8-2} and \textbf{T19-1,2} use \textit{WavLM} features.

\item
\textit{Phonetic transcriptions}:
\textbf{T12-2,3,4}, \textbf{T30-2} employ cascaded ASR systems (\textit{Whisper}, hybrid CTC-Attention ASR) to obtain phonetic transcriptions 
or intermediate representations that are then fed to TTS systems to encourage the preservation of linguistic information at the transcription level.

\end{itemize}

\section{Results}
\label{sec:results}

\begin{figure}
    \centering
    \includegraphics[width=1.02\linewidth]{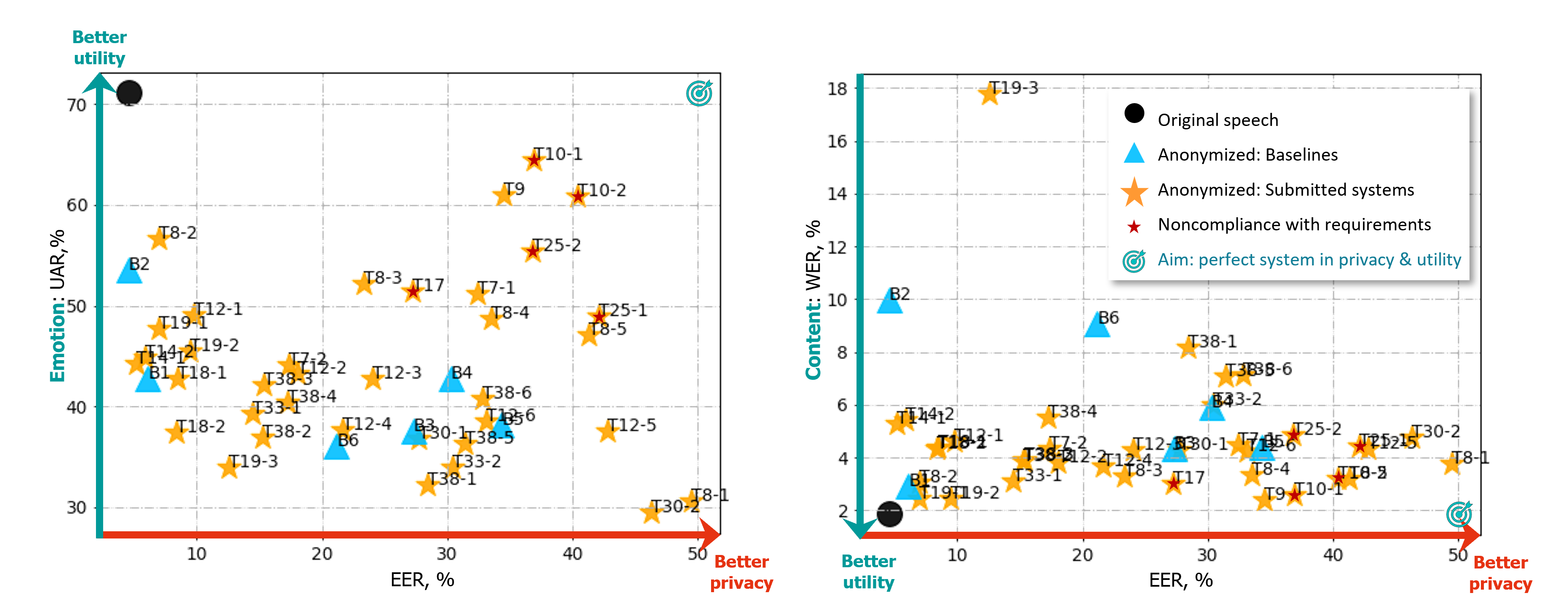}
    \caption{Challenge results on the test data}
    \label{fig:results}
\end{figure}

A summary of results in terms of utility (WER,\% or UAR,\%) against privacy (EER,\%) is illustrated in Figure \ref{fig:results} for
submitted systems (orange stars) described in Section \ref{sec:participants_anonym_systems} and baseline systems (blue triangles) described in Section \ref{sec:baseline}. 
For all cases, EER results are derived using ASV models re-trained by the organizers using anonymized data provided by participants.
Results for original, unprotected data are indicated with black circles. 
Each EER estimate is an average of EERs derived separately using male and female speaker subsets computed using the \textit{LibriSpeech} \textit{test-clean} dataset. 
Detailed results for development and test datasets can be found in Table~\ref{tab:results}.

\subsection{Privacy achievement across categories}

\begin{figure}[h]
    \centering
    \includegraphics[width=0.9\linewidth]{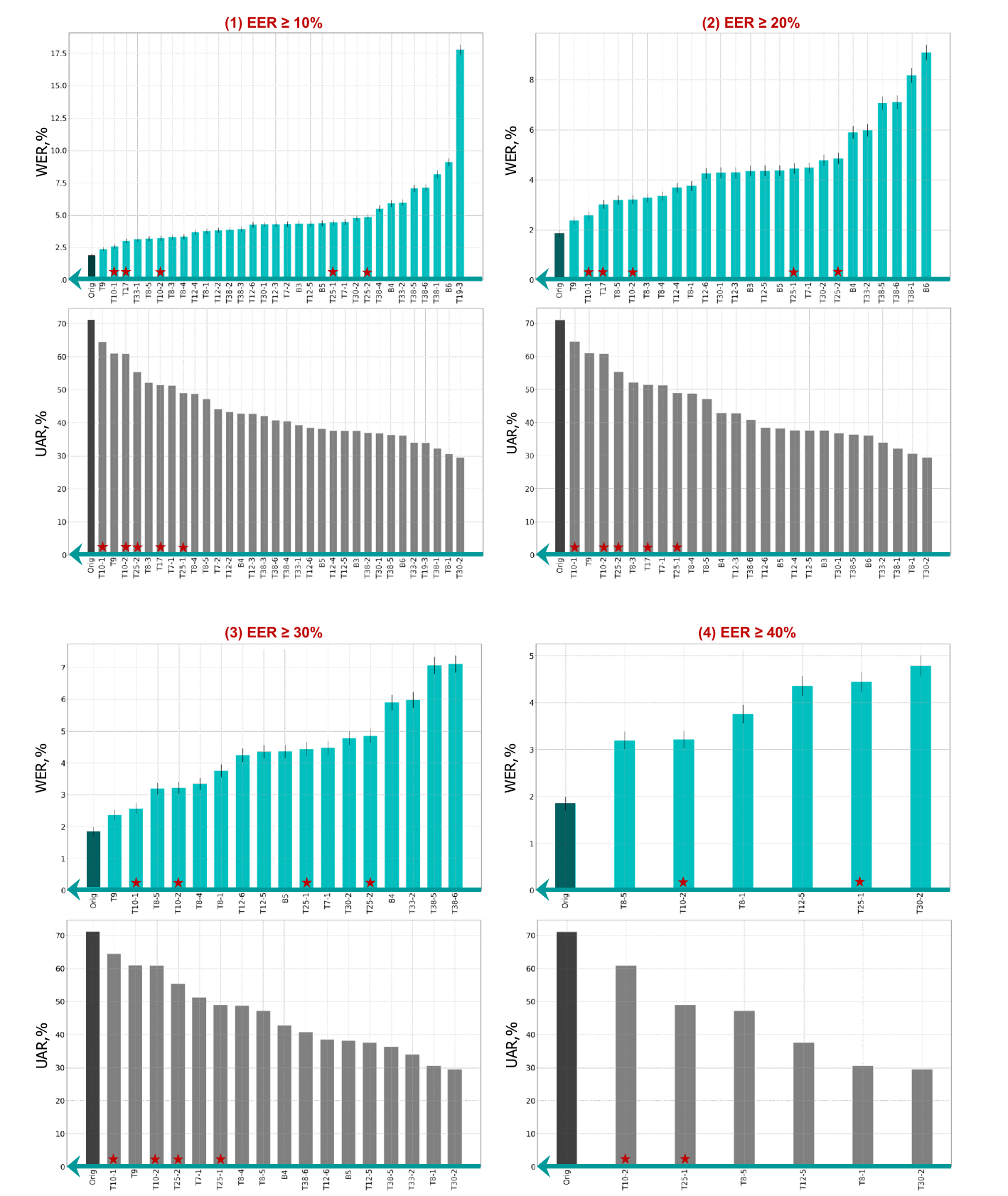}
    \caption{Anonymization system ranking according to the utility  metrics (WER,\% and UAR,\%) for 4  privacy thresholds on the test data (official ranking).}
    \label{fig:ranking}
\end{figure}

Results and rankings for utility metrics (WER,\% and UAR,\%) against the EER are illustrated in Figure \ref{fig:ranking}
across four different privacy thresholds (10, 20, 30, and 40\% of EER).
WER values are given with $95\%$ confidence intervals.
The EER for original, unprotected test data is 4.6\%.
For the first category (EER $\geq10\%$) there are 28 submitted systems and 4 baselines. 
Of these, 21 submitted systems and all 4 baselines meet the requirements of  the second category (EER $\geq20\%$). 
For the third category (EER $\geq30\%$) there are 15 submitted systems and 2 baselines,
while only 6 submitted systems (\textbf{T10-2}, \textbf{T8-5}, \textbf{T25-1}, \textbf{T12-5}, \textbf{T30-2}, and \textbf{T8-1}) meet the criteria for the fourth strongest category (EER $>40\%$). However, two of these (\textbf{T10-2} and \textbf{T12-5}) are non-compliant with the challenge rules.
The highest EER, close to 50\%, is achieved by the cascaded ASR-TTS system \textbf{T8-1}.

\subsection{Comparative performance by system type}

\subsubsection{Neural codec-based systems (T10, T12-1, T17)}
Figure~\ref{fig:ranking} demonstrates that 
some neural codec systems achieve 
% exceptional 
strong
performance in the highest privacy category. 
\textbf{T10-2} (EER=40.4\%), which combines a codec with serial disentanglement and frame-level emotion distillation, achieves the best overall trade-off between privacy protection (EER $\geq40\%$) and preservation of both linguistic content (WER=3.2\%)  and emotional expressiveness (UAR=60.8\%). 
\textbf{T17}, a FACodec-based system, is in the second category (EER=27\%) and demonstrates is also effective in preserving utility WER=3\%, UAR=51.4\%).
\textbf{T12-1}, which leverages a FACodec with pool-based averaging, cross-gender conversion, and AWGN, 
offers weaker privacy protection (EER=9.7\%) and
 moderate  utility characteristics.

\subsubsection{Cascaded ASR+TTS systems (T8-1, T12-2,3,4, T30-2)}

These systems achieve near-perfect speaker anonymization with EER values for one  (\textbf{T8-1}) approaching 50\%. However, as seen in Figure~\ref{fig:results},  cascaded systems systematically show substantially lower UAR values compared to neural codec and VC-based alternatives.
\textbf{T8-1}, a cascaded ASR-TTS system using \textit{Whisper-VITS}, achieves the strongest privacy (EER=49.5\%) but is among the worst for emotion preservation (UAR=30.6\%).
However, linguistic content preservation remains strong (WER=3.75\%, the third-ranked system in the highest privacy category).
The asymmetric preservation of content through the transcription bottleneck but degradation of paralinguistic attributes highlights the main limitations of text-based anonymization.

Among cascaded ASR-TTS systems with paralinguistic preservation, 
\textbf{T30-2} in the same highest category (EER=46.3\%), achieves worse results than \textbf{T8-1} for both utility metrics (WER=4.78\%, UAR=29.4\%), indicating that  the incorporation of latent features based on original F0 and emotion prototype vectors may lead to a minor reduction in privacy while not improving emotion preservation.
\textbf{T12-2,3,4} -- \textbf{B3}-based anonymization systems -- fall into the first (\textbf{T12-2}) and second (\textbf{T12-3,4}) privacy categories. 
\textbf{T12-2} and \textbf{T12-3} use emotion encoders that improve UAR (up to 43\%), which is the best result for this type of system. For comparison, similar systems without emotion encoders (i.e., \textbf{T12-4, B3}) achieve UAR values not exceeding 38\%.

\subsubsection{Hybrid admixture systems (T8-3,4,5)}

\textbf{T8}'s \textit{admixture} approaches, which randomly select between cascaded ASR+TTS (\textbf{T8-1}) and kNN-VC (\textbf{T8-2}) on a per-utterance basis, achieve above-average performance across privacy-utility objectives. 
The mixing probability allows flexible adjustment of the privacy-utility tradeoff within each privacy category: UAR changes proportionally to EER, while WER remains similar across all systems (approximately 3.2--3.3\%). 
\textbf{T8-5}, which ranks among the top systems, achieves EER=41.3\% while maintaining UAR closer to the levels achieved by pure VC systems (UAR=47\%), demonstrating that hybrid approaches effectively enable context-dependent privacy-utility optimization.

\subsubsection{VC-based systems (T7, T8-2, T9, T12-5,6, T14, T18, T19-1,2,3, T25, T33, T38)}

Pure VC approaches generally struggle to achieve the highest privacy categories but in some cases show good utility preservation.
kNN-VC variants (\textbf{T8-2}, \textbf{T19-1,2}) achieve relatively strong WER and UAR values but typically their EER does not exceed 10\%.
Emotion-aware VC systems such as \textbf{T25-1,2} and \textbf{T9} demonstrate improved emotion preservation compared to most baseline VC systems but remain below neural codec performance (e.g., \textbf{T10-1,2}) in the highest privacy categories.
\textbf{T9} ranks among the top systems in the first three privacy categories (EER=34.5\%, WER=2.4\%, UAR=61\%).

\subsection{Utility metrics analysis}

\subsubsection{Linguistic content preservation (WER)}

The WER on unprocessed test data is 1.85\%. 
In the first three privacy categories, the best system for content preservation, \textbf{T9}, achieves WER=2.37\% on the test data, corresponding to 0.5\% absolute and 28\% relative WER increase. 
In the highest privacy category, the two systems with the lowest WER (3.2\%) -- the cascaded system \textbf{T8-5} and the codec-based system \textbf{T10-2} -- demonstrate more noticeable ASR performance degradation: approximately 1.4\% absolute and 73--74\% relative. 
Within the highest privacy category, WER ranges from 3.2\% to 4.8\%.

\subsubsection{Emotion preservation (UAR)}

The UAR on the original (unprocessed) test data, marked with a black circle in Figure~\ref{fig:results}, is 71\%. In the first three privacy categories, the top systems (\textbf{T10-1,2}, \textbf{T9}) achieve UAR values of approximately 61\%–65\%, while rule-compliant systems (\textbf{T9}, \textbf{T7-1}, \textbf{T8-4}) achieve approximately 49\%–61\% in the third privacy category and 51\%–61\% (\textbf{T9}, \textbf{T8-3}, \textbf{T7-1}) in the first two privacy categories. In the highest privacy category, top-performing systems (\textbf{T10-2}, \textbf{T25-1}, \textbf{T8-5}) achieve UAR values of approximately 47\%–61\%, and the best rule-compliant system (\textbf{T8-5}) achieves 47\%. The spread between best and worst rule-compliant systems (\textbf{T8-5} and \textbf{T30-2}) in the highest privacy category exceeds 17\% (and 31\% for all submitted systems in this category) in UAR, underscoring substantial performance differences and highlighting the importance of emotion-aware architecture choices. Systems explicitly incorporating emotion preservation mechanisms (\textbf{T9}, \textbf{T10-1,2}, \textbf{T25-1,2}) rank higher in UAR across all privacy categories compared to systems without emotion-specific designs.

Top systems (\textbf{T9}, \textbf{T10}, \textbf{T25-2})
achieving high privacy categories (EER$\geq$30\%,40\%) also rank among the best performers in lower privacy categories (EER$\geq$10\%,20\%) for emotion preservation. 
This consistency across privacy thresholds indicates that these systems benefit from principled architectural designs, such as multi-stage emotion-aware disentanglement, emotion-matched pseudo-speaker selection, and hybrid anonymization approaches, that do not trade off emotion preservation for privacy gains. 

\begin{figure}[h]
    \centering
    \includegraphics[width=0.82\linewidth]{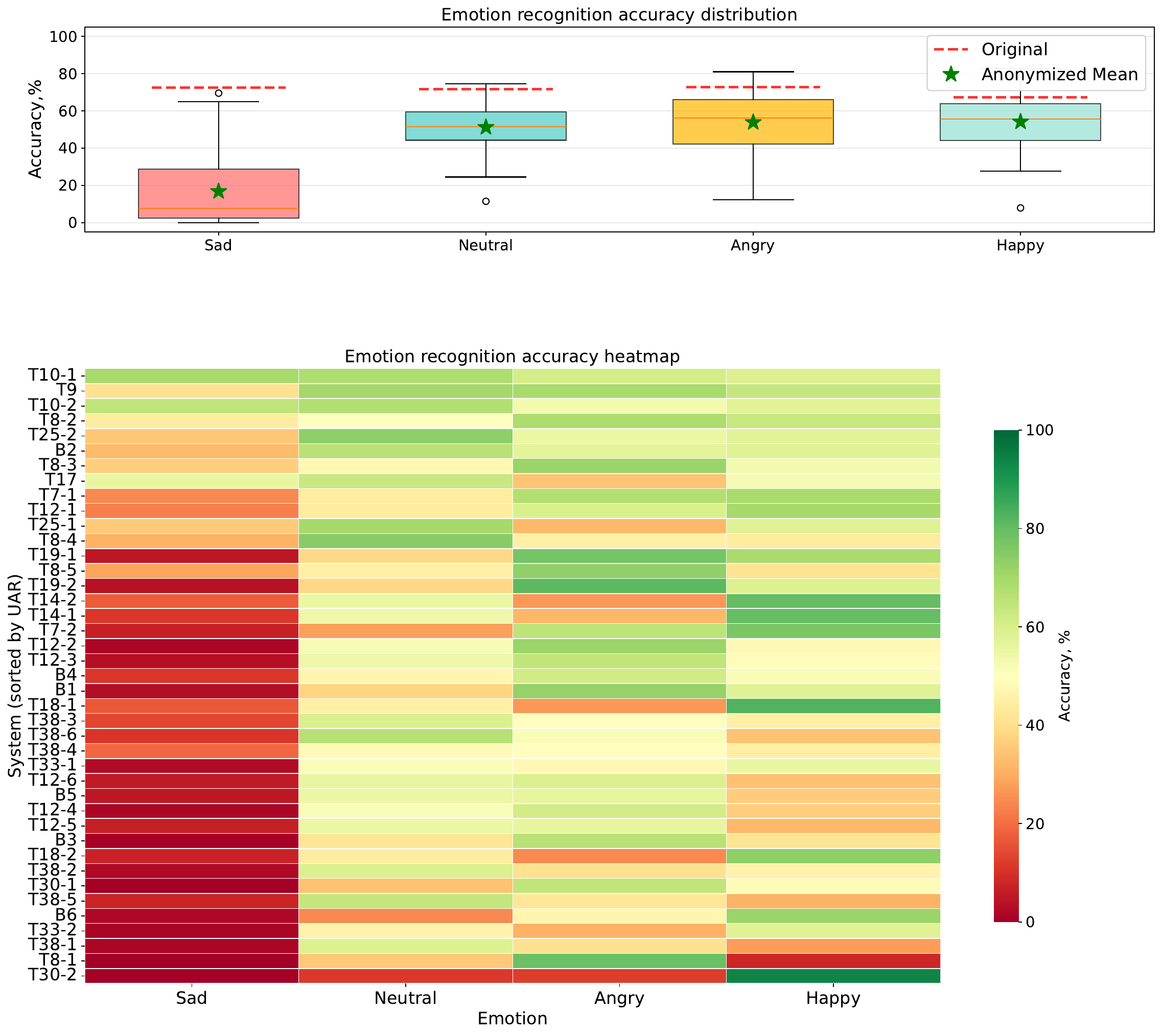}
    \caption{Emotion recognition accuracy by emotion class on the \textit{IEMOCAP} test set}
    \label{fig:emotion}
\end{figure}

Figure~\ref{fig:emotion} shows emotion-specific accuracy. For original speech, accuracy is balanced across all four classes. Anonymization impacts accuracy differently for specific emotion classes. In particular, many systems fail to recognize \textit{sadness}, which is the hardest class to preserve and has the lowest accuracy across all anonymized systems. Balanced systems include \textbf{T10-1}, which preserves all emotions well (60–70\% accuracy for each). \textit{Happiness}-biased systems include \textbf{T18-1}, \textbf{T14-1,2}, and \textbf{T30-2}, which achieve excellent \textit{happiness} accuracy (80–94\%) but poor \textit{sadness} accuracy (29–45\%). Systems \textbf{T8-1}, \textbf{T30-1}, \textbf{T30-2}, \textbf{B3}, and \textbf{T33-2} have near 0\% accuracy for \textit{sadness}.

\subsection{Privacy-utility trade-off}

The VPC 2024 reveals fundamental trade-offs across all evaluation metrics, with no single system simultaneously maximizing privacy (EER), linguistic content preservation (WER), and emotion preservation (UAR). The challenge results demonstrate a non-uniform operating space where the relationship between privacy and utility varies substantially across different metrics and operating points.

\begin{figure}[h]
    \centering
\includegraphics[width=0.95\linewidth]{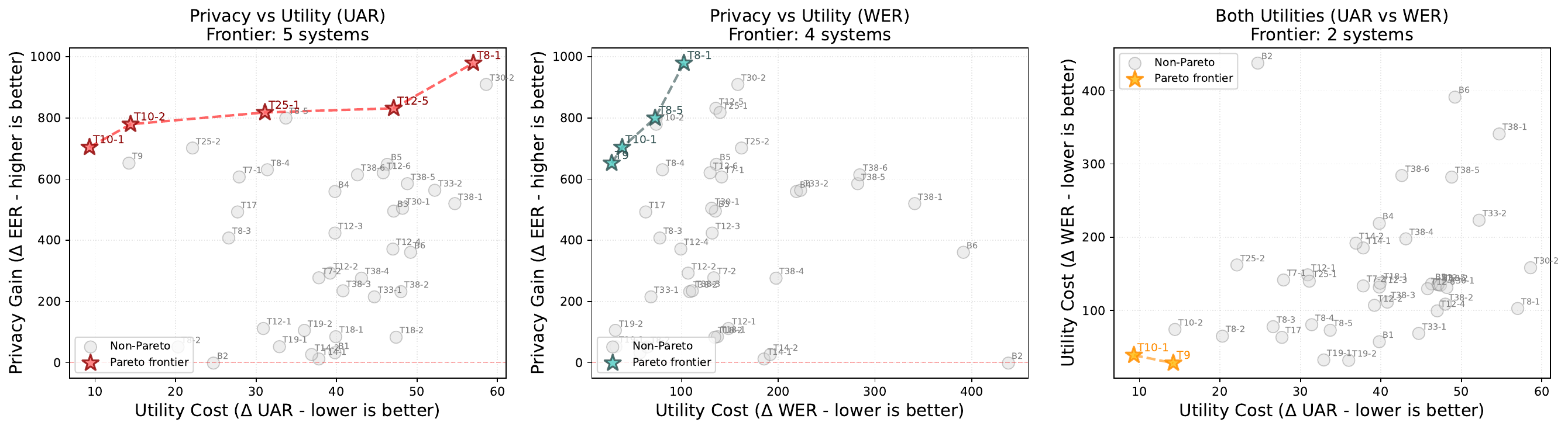}
    \caption{Pareto curve analysis on the test sets}
    \label{fig:pareto}
\end{figure}

 Figure~\ref{fig:pareto} presents scatterplots of relative metric changes ($\Delta$WER, $\Delta$UAR, and $\Delta$EER) for anonymization systems with respect to results on original data. The Pareto frontiers shown across these plots reveal distinct trade-off patterns.
The first plot (\textit{Privacy vs. Utility (UAR)}) identifies systems on the frontier, with \textbf{T10-1} achieving the best balance ($\Delta$UAR = 9\%, $\Delta$EER = 704\%),
followed by \textbf{T10-2} ($\Delta$UAR=14\%, $\Delta$EER = 779\%) and \textbf{T25-1} ($\Delta$UAR=31\%, $\Delta$EER=818\%). 
In contrast, \textbf{T8-1} achieves the strongest privacy ($\Delta$EER = 979\%) but at significant utility cost ($\Delta$UAR=57\%). 
The second plot (\textit{Privacy vs. Utility (WER)}) demonstrates that \textbf{T9} achieves a good trade-off ($\Delta$WER = 28\%, $\Delta$EER = 652\%), while \textbf{T10-1} again performs competitively ($\Delta$WER = 39\%, $\Delta$EER = 704\%).
The third plot (\textit{Both Utilities (UAR vs. WER)}) identifies systems that preserve both utility metrics simultaneously, with \textbf{T9} ($\Delta$UAR = 14\%, $\Delta$WER = 28\%) and \textbf{T10-1} ($\Delta$UAR = 9\%, $\Delta$WER = 39\%) representing the most effective approaches. These Pareto frontiers underscore the critical role of system design choices in achieving balanced privacy-utility trade-offs.

Figure~\ref{fig:heatmap-all} presents a normalized performance matrix displaying all systems ranked by overall combined score, enabling direct comparison of privacy-utility trade-offs across all submission approaches.
The challenge's multi-threshold evaluation shows that optimal system selection depends on application requirements. 
\textit{Admixture} systems (\textbf{T8-3,4,5}) demonstrate that dynamic method selection on a per-utterance basis enables flexible operating points, with mixing probability controlling positions along the privacy-utility spectrum.

\subsection{Gender impact on anonymization}

\begin{figure}[h]
    \centering
    \includegraphics[width=0.85\linewidth]{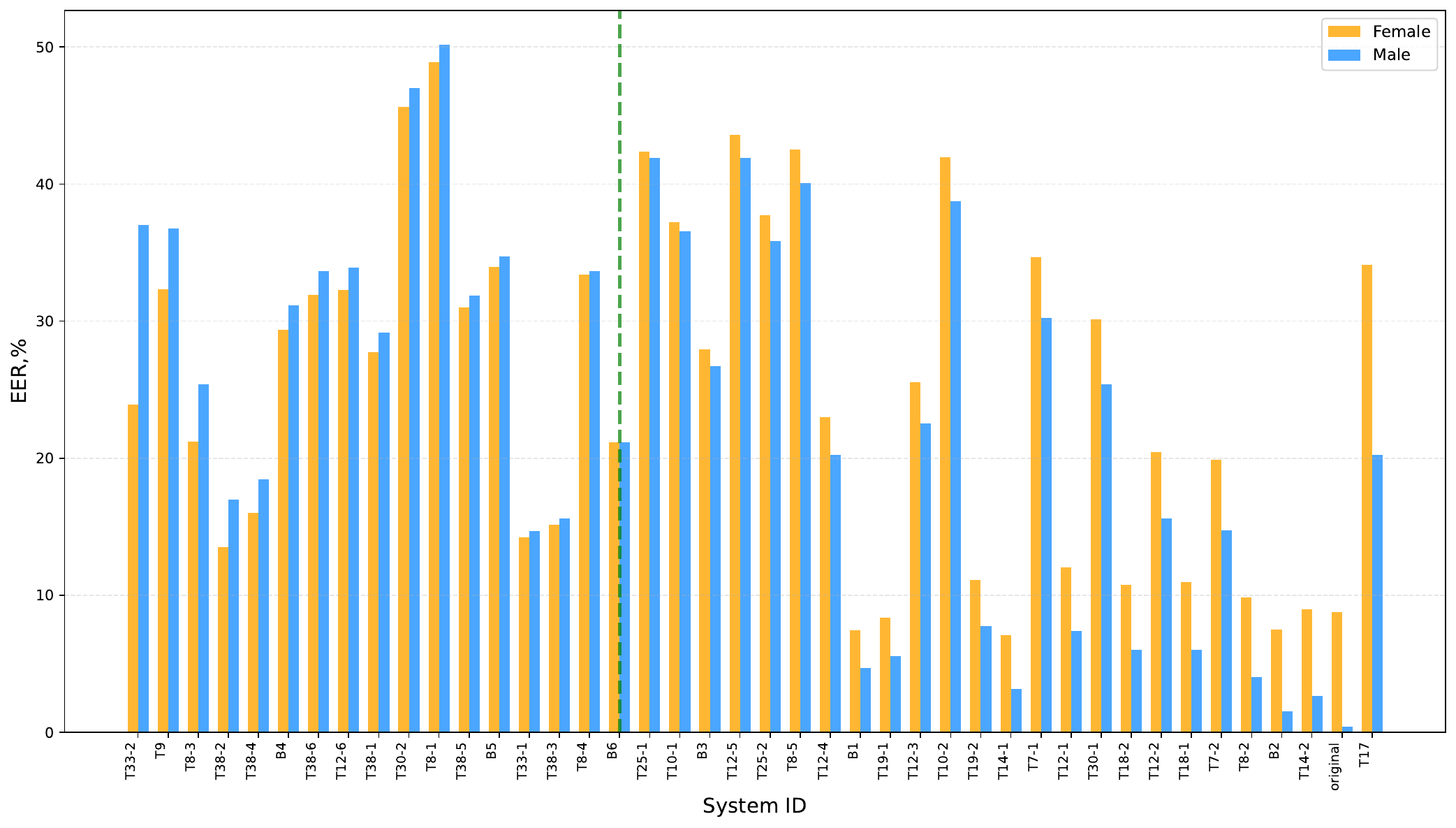}
    \caption{Gender bias in anonymization: systems ordered by male-female EER  difference}
    \label{fig:gender-eer}
\end{figure}

Figure~\ref{fig:gender-eer} displays EER results for the test set for female and male speaker subsets, ordered by the difference ($\text{EER}_\text{male}-\text{EER}_\text{female}$) in descending order.
Gender-based performance disparities vary significantly across systems. The original ASV system on unprotected data exhibits substantial male bias (EER=8.8\% for female speakers and EER=0.4\% for male speakers), indicating that male speaker verification is inherently easier.
The most gender-biased anonymisation systems include \textbf{T17}, which exhibits a 14\% absolute difference ($\text{EER}_\text{female}$=34\%, $\text{EER}_\text{male}$=20\%), and \textbf{T33-2}, which shows females bias ($\text{EER}_\text{female}$=24\%, $\text{EER}_\text{male}$=37\%).
Gender-balanced systems include \textbf{B6}, \textbf{T8-4}, \textbf{T25-1}, \textbf{T38-3}, and \textbf{T33-1}, which maintain differences below 0.5\%, demonstrating uniform anonymization effectiveness across genders.

\begin{figure}[h]
    \centering
    \includegraphics[width=1\linewidth]{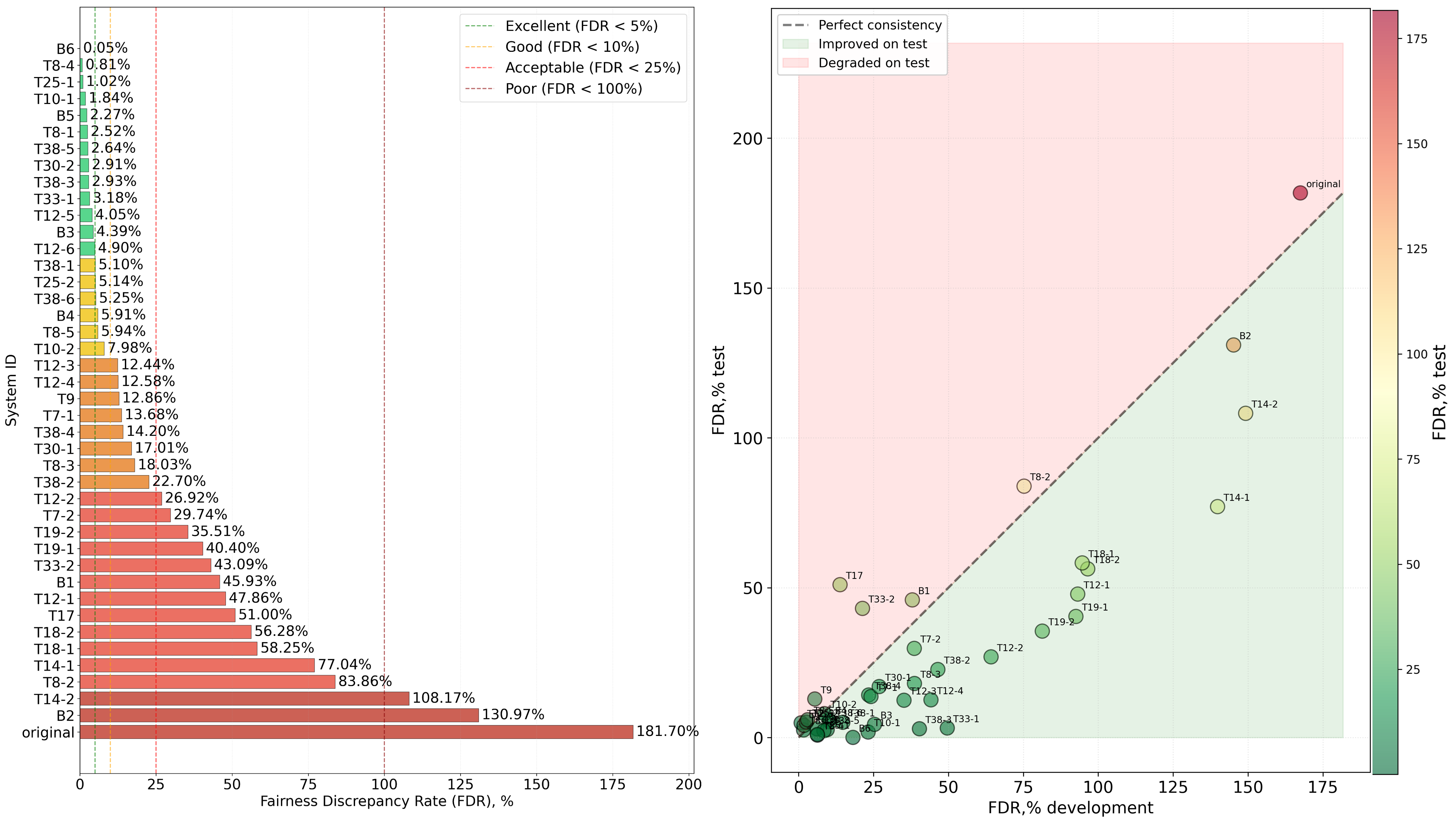}
    \caption{Gender fairness  analysis in anonymization: (1) System ranking according mFDR results on the test set; (2)~mFDR consistency across development and test sets.}
    \label{fig:gender-fdr}
\end{figure}

To further assess gender fairness in anonymization, we employ the \textit{modified} \textit{fairness discrepancy rate} (mFDR), evaluated at the EER operating point, which captures the relative magnitude of performance difference between genders. This simplified metric, adapted for speaker verification systems and VPC 2024 evaluation conditions, is defined as follows~\footnote{A similar approach to assess gender impact in the speech domain was used in \citep{zanon2022study}.
A more general definition of fairness discrepancy rate (FDR) in biometric systems can be found in~\citep{de2021fairness}, which is calculated for operational conditions where a single threshold is set for all demographic groups. 
Such a metric, in application to the VPC 2024 evaluation setup (EER operational point, two gender groups), 
would correspond to a weighted sum of the absolute differences in FRR and FAR between female and male speakers.
The mFDR, while using normalization of the gap by the overall performance level, addresses issues of difference-based bias measures that lose sensitivity when base error rates are small or differ in magnitude across conditions. From this perspective, mFDR is closer in spirit to \textit{Gini aggregation rate for biometric equitability} (GARBE)~\citep{howard2022evaluating,chouchane2024comparison}.}:

\begin{equation*}
    \text{mFDR} =  \frac{2\cdot |\text{EER}_\text{female} - \text{EER}_\text{male}|}{(\text{EER}_\text{female} + \text{EER}_\text{male})} \cdot 100.
\end{equation*}

Figure~\ref{fig:gender-fdr} displays gender fairness analysis for all anonymization systems using the mFDR metric on both test and development datasets. Systems are categorized by mFDR performance, ranging from \textit{excellent} (mFDR$<$5\%) for systems such as \textbf{B6}, \textbf{T8-4}, \textbf{T25-1}, \textbf{T10-1}, \textbf{B5}, \textbf{T8-1}, and \textbf{T38-5} to \textit{poor} for systems \textbf{B2}, \textbf{T14-2}, \textbf{T8-2}, and \textbf{T14-1}. The maximum mFDR value is observed on the original data (182\%).
Figure~\ref{fig:gender-fdr} (2) shows consistency of mFDR across the development and test sets. For most of the anonymization systems, mFDR improves on the test set.

\subsection{Baseline influence}

Table~\ref{tab:summary_s} presents the baseline influence on submitted systems. The baseline systems (\textbf{B1}–\textbf{B6}) established performance boundaries that participant systems aimed to improve upon. The x-vector anonymization strategy based on pool selection and averaging, introduced in \textbf{B1}, was employed in \textbf{T14-1} and \textbf{T14-2}.
Multiple cascaded ASR-TTS systems with paralinguistic preservation are based on \textbf{B3}, which uses phonetic transcriptions and GAN-based pseudo-speaker generation. \textbf{T12-2}, \textbf{T12-3}, \textbf{T12-4}, and \textbf{T30-1} build directly upon \textbf{B3}, introducing emotion awareness and alternative anonymization strategies. Performance for these systems is similar to \textbf{B3}, with slight improvements in overall utility for \textbf{T12-2} and \textbf{T12-3}.
\textbf{T12-5} and \textbf{T12-6} modify \textbf{B5} (ASR-BN with VQ-based content features) by replacing the pitch curve with a weighted average between the original curve and its moving average. \textbf{T12-6} additionally applies white Gaussian noise to the resulting pitch curve. These systems achieve performance close to \textbf{B5}. In contrast, \textbf{T25-1} and \textbf{T25-2} employ VQ-BN for content features while introducing emotion transfer mechanisms, achieving significantly better privacy and UAR metrics than \textbf{B5}. VQ-based approaches are also employed in \textbf{T19-3}.
The success of neural codec baseline \textbf{B4} motivated substantial community adoption, with \textbf{T10}, \textbf{T12-1}, and \textbf{T17} representing independent neural codec approaches that achieve competitive performance. \textbf{T10} and \textbf{T17} significantly outperform \textbf{B4} in both WER and UAR, and \textbf{T10} systems also achieve higher EER values.
Overall, the best submitted systems significantly outperform baselines across all metrics. EER improves from 34\% (\textbf{B5}) to 49.5\% (\textbf{T8-1}), UAR improves from 54\% (\textbf{B2}) to 65\% (\textbf{T10-1}), and WER improves from 2.9\% (\textbf{B1}) to 2.37\% (\textbf{T9}).

\subsection{Summary}

Cascaded approaches clearly dominate in absolute privacy (EER = 48\%) but fail in emotion preservation. Neural codec methods achieve the best overall balance, with \textbf{T10-2} representing state-of-the-art privacy-utility compromise. Hybrid approaches successfully bridge different optimization targets through dynamic method selection. Advanced VC systems with explicit attribute disentanglement and self-supervised learning (SSL) representations demonstrate substantially improved performance compared to traditional VC, reflecting the field's adoption of deep learning advances.
The challenge results demonstrate that while strong privacy protection remains achievable, it requires sophisticated architectural choices that balance multiple competing objectives. The evolution from simple baseline approaches to complex multi-stage disentanglement systems reflects rapid methodological development in voice anonymization research.

\section{Discussion and future perspectives}
 \label{sec:Discussion}

In this section we present key insights from the VPC 2024 and previous editions, examine evaluation methodology and data limitations, survey emerging research directions, and provide strategic recommendations for future voice anonymization research.

\subsection{Evaluation}

Reliable evaluation is key to the fostering of progress in anonymization while also helping to ensure real-world applicability. However, current evaluation methodologies present limitations that require careful analysis.

\subsubsection{Reliability of privacy evaluation}  
\label{sec:attacker_challenge} 

EERs approaching 50\% under the \textit{semi-informed} attack model (see Section~\ref{subsec:attack_model}) indicate ideal privacy protection. 
Many systems submitted to the VPC appear to achieve such high levels of privacy.
However, several recent analyses, call into question whether such encouraging estimates of anonymization performance are reliable.
Results from the \textbf{First VoicePrivacy Attacker Challenge}\footnote{\url{https://www.voiceprivacychallenge.org/attacker/}}~\citep{tomashenko2025first} and other works~\citep{panariello2025risks} show  that exaggerated EERs can be the result of poor attacker generalization rather than strong anonymization.
Meanwhile, the current attack model imposes a number of constraints which can also contribute to the overestimation of privacy.  These include:
\begin{itemize}
\item \textit{Data}: the attacker 
trains an ASV system using anonymized data produced by each system. In more realistic scenarios, attackers may access arbitrary external corpora, including other sets of anonymized speech, to strengthen the attack.
\item \textit{Features}: the attacker uses Fbank input features and trains a speaker encoder following a typical ASV pipeline, while other studies~\citep{tomashenko2025exploiting,tomashenko2025analysis,bakari25_spsc} show that context-dependent duration features, temporal speech dynamics, and non-timbral cues can all effectively reveal speaker-specific information after anonymization.
\item \textit{Model and hyperparameters}: attacks are performed using a fixed architecture (i.e. ECAPA-TDNN) and untuned hyperparameters, whereas there is evidence that stronger attacks can be implemented using other models~\citep{bakari25_spsc}, e.g. \textit{WavLM}.

\end{itemize}

These limitations were  addressed in the First VoicePrivacy Attacker Challenge \citep{tomashenko2025first,tomashenko2024first},  an ICASSP 2025 SP Grand Challenge which evaluated attacks against a selection of voice anonymization systems.
The use of strong attacks is crucial to ensure reliable privacy evaluation.
Seven anonymization systems were considered. 
They were the best three VPC 2024 baselines, namely \textbf{B3}, \textbf{B4} and \textbf{B5} (see  Section~\ref{sec:baseline}), and a selection of four participant systems which (initially) achieved the highest level of privacy (EER$\geq$40\%, category~4 in Figure~\ref{fig:thresholds}) when evaluated using the baseline attack model, namely \textbf{T8-5}, \textbf{T10-2}, \textbf{T12-5}, and \textbf{T25-1}.  
When evaluated with the baseline attacker, these systems achieved the 4th privacy category (EER$\geq$40\%). 
However, when tested against the best attack models developed by Attacker Challenge participants, EERs for all systems fall substantially  
(EER$\geq$20\%, category~2). 
Results also show that system ranking based on EERs  changes when evaluation is performed using stronger attack models.
These finding highlight the importance to reliable and trustworthy privacy evaluation of using the strongest possible attack models.

\subsubsection{Metrics} 

Since the 2020 VPC, a wide range of evaluation metrics have been proposed, covering both privacy and utility aspects.
Nevertheless, several limitations and concerns remain:

\paragraph{\textbf{Privacy metrics.}} 
  Use of the EER fails to reflect legal privacy requirements such as \textit{linkability}, \textit{singling out}, or \textit{inference}~\citep{Article29_2014}.
The EER can be insensitive to re-identification risks, remaining stable even as the probability of successful re-identification varies for specific users or use cases. 
It focuses only on average performance at a single operating point, which masks worst-case risks and outlier vulnerabilities, potentially giving an overly optimistic view of privacy protection in the case of individuals who remain readily identifiable even after anonymisation~\citep{nautsch2020zebra,williams2024anonymizing}.

To better align with both technical and legal expectations, alternative evaluation frameworks have been proposed. 
One such approach, based on \textit{linkability} and \textit{singling out}, and inspired by the legally accepted interpretation of the definitions of anonymous information was proposed in~\citet{vauquier25_interspeech}.
Results derived using these metrics indicate that residual privacy risks after anonymization are more substantial than suggested by EER estimates, underscoring the need for evaluation metrics that are better aligned with both technical and legal perspectives.
Alternative metrics also include 
the \textit{similarity rank disclosure} (SRD) metric proposed by \cite{backstrom2025privacy}, measured in units of entropy (bits). It is defined as the reduction in entropy (uncertainty) about the true identity after observing the similarity rank.
The  metric quantifies how much personally identifiable information (PII) about the true identity can be inferred from the similarity rank when comparing a query sample to a database of templates and shows promise as a metric for the evaluation of privacy in voice anonymisation~\citep{chandraPetteno25}.
The adoption in future VPC editions of metrics that reflect performance for the most challenging or privacy-sensitive samples might provide estimates of anonymization performance that are better aligned with expectations in real-world settings.

\paragraph{\textbf{Utility metrics.}}

One limitation of current utility metrics is the disconnect between automatic metrics and perceptual quality. While more powerful ASR models generally yield lower WERs, this does not necessarily indicate improved quality, or even intelligibility. Studies demonstrate that WER estimates are not always 
positively 
correlated with subjective intelligibility~\citep{VPC2024}, highlighting a fundamental gap between automatic evaluation and human perception.
Additionally, while estimates of the WER reflect the preservation of speech content, they neglect other important aspects of utility. Speech naturalness, prosody preservation, and acoustic quality are underrepresented by current metrics.

Proposed solutions include developing multi-dimensional metrics that assess both objective and subjective utility, 
integrating evaluation criteria for specific use cases, 
and designing  utility metrics that jointly evaluate  intelligibility, naturalness, and downstream task utility.

\subsubsection{Data and benchmarking}

All previous VPC editions have relied primarily on the \textit{LibriSpeech} datasets for the evaluation of privacy and content preservation, with VCTK also having been used for the 2020 and 2022 editions. These corpora contain clean, single-speaker, read English speech. Since the latter lack emotional annotations, the \textit{IEMOCAP} corpus, which features spontaneous and emotionally expressive speech, was introduced for the 2024 edition to support the evaluation of emotion preservation. However, due to the limited number of speakers and size, \textit{IEMOCAP} is itself not suited to the evaluation of privacy and content preservation.

There are some limitations of the current datasets and experimental protocols:
(1)~\textit{Dataset inconsistency in privacy and utility evaluation.} 
The estimation of privacy and content preservation from one dataset, and emotion preservation from another dataset might introduce some bias in the results, raisings concerns about whether comparisons of system performance are fair.
(2)~\textit{Language and accent coverage.} Speech data are of English language, with minimal representation of multilingual, code-switched, or heavily accented speech.
(3)~\textit{Speech characteristics.} \textit{LibriSpeech} contains only clean, read speech from a limited demographic; the dataset lacks spontaneous speech, child speech, pathological speech, and speaker variations reflecting real-world diversity.
(4)~\textit{Recording conditions.} There is no systematic evaluation of anonymization robustness under noisy, reverberant, or compressed acoustic conditions typical of many real-world applications (e.g., phone calls, live streaming, video conferencing).
(5)~\textit{Conversational scenarios.} \textit{LibriSpeech} and \textit{IEMOCAP} are single-speaker,
and current solutions explored in the challenge are limited to single-speaker anonymization. These methods are not yet directly applicable to multi-speaker or overlapping speech scenarios~\cite{miao2024benchmark, tomashenko2025target}.

Future voice anonymization research demands datasets that are not only larger but also far more diverse, encompassing data collected from speakers of different ages, accents, languages, speech styles, emotional states, and acoustic conditions. 
To ensure anonymization techniques are robust and universally applicable, there is also a need for datasets which reflect more challenging and realistic use-case scenarios, such as those involving overlapping speech and low-resource languages; 
and
establishing fairness benchmarks with assessment whether anonymization performance is consistent across speaker demographics (age, gender, accent, language background, speech disorders).

\subsection{Emerging and advanced approaches to anonymization}

In this section we survey recent advances in  anonymization 
highlighting both methodological innovations and application-specific developments.
The majority of recent work continues to build upon disentanglement-based methods, which explicitly decompose speech into independent factors to enable controlled anonymization. Several  primary approaches have emerged:

% \item 
\textit{Multi-attribute disentanglement}  extends single-attribute approaches to enable simultaneous disentanglement of multiple speech attributes~\citep{miao2025adapting,yao25_interspeech}. This advancement enables more fine-grained, context-aware anonymization that can simultaneously control voice identity, emotional expression, demographic attributes, and other speaker characteristics, offering greater flexibility for domain-specific applications.

\textit{Speaker-embedding-free approaches}
~\citep{cai2024privacy,das2025,tang2025sef}  avoid explicit speaker embeddings entirely. Instead, these approaches learn anonymized representations directly from speech signals, potentially simplifying the anonymization pipeline and improving generalization to unseen acoustic conditions and speaker populations. 
 
Recent work has addressed other, diverse applications and challenging scenarios. It is organized along three primary dimensions: deployment constraints, target populations, and paradigm variants:

\textit{Language and cross-lingual generalization}:
Emerging work in multilingual anonymization~\citep{miao2022language,miao2022analyzing,deng2023v,miao2023language,meyer2024probing,yao2024musa} and code-switching scenarios~\citep{meyer2025first} shows that anonymization models can generalize to other languages and mixed-language speech. 
Cross-lingual generalization broadens real-world applicability 
without the need for multilingual training data or any retraining/adaption.

\textit{Conversational scenarios and multi-speaker recordings}:
Many natural use cases for anonymisation involve multiple, concurrent speakers, 
and complex conversational dynamics that complicate anonymization~\citep{miao2024benchmark}.
Recent work~\citep{tomashenko2025target} addresses the challenge of anonymizing a single target speaker in multi-speaker overlapping conversational recordings, and proposes new methods and improved evaluation strategies to enhance privacy protection and utility assessment in these complex scenarios.

\textit{Recording conditions and noisy speech}:
Only few works address the evaluation of anonymization robustness under noisy, reverberant, or compressed acoustic conditions which typify many real-world applications (e.g., phone calls, live streaming, video conferencing). 
Work in~\cite{miao2024synvox2}  shows that the application of anonymization techniques to noisy, spontaneous speech from the \textit{VoxCeleb-2} dataset can distort speech content.

\textit{Real-time and low-latency anonymization}: is critical for live applications such as streaming, video conferencing, and telephony. Recent work~\citep{quamer2024end} addresses this challenge by replacing computationally intensive models (e.g., \textit{HuBERT}, HiFi-GAN) with lightweight convolutional neural network architectures and chunk-wise speech processing, achieving anonymization under strict latency constraints without substantial utility degradation.

\textit{Target populations and specialized speech domains}:
Recent work has extended anonymization techniques to underrepresented populations and specialized speech domains.
Techniques tailored to children's voices should address unique acoustic characteristics and developmental variations~\citep{kulkarni25b_interspeech}.
Methods for pathological speech (dysarthria, dysphonia, etc.) explored by~\citet{tayebi2024addressing} maintain intelligibility and clinical diagnostic utility while protecting voice identity.

\textit{Synchronous and asynchronous anonymization}:
Two complementary anonymization paradigms have emerged, each with distinct use cases~\citep{chen2025any}.
\textit{Synchronous} anonymization modifies both machine-perceived and human-perceived voice identity, altering speech signals so that both automatic systems and human listeners perceive a different voice identity. This approach aligns with the VPC definition of voice anonymization. 
\textit{Asynchronous} anonymization modifies only machine-discernible speaker attributes while preserving human perception of the original voice identity.

\textit{User control and flexibility}:
Recent work~\citep{hui2025securespeech} introduces prompt-based speech anonymization leveraging large language model (LLM)-based ASR-TTS frameworks. This approach enables flexible, user-controllable anonymization through natural language prompts that specify desired speaker attributes, while simultaneously addressing content privacy through named entity recognition and substitution. 
This framework opens new possibilities for context-dependent and task-specific anonymization strategies.

\textit{Beyond voice identity: multi-attribute privacy protection}:
In addition to voice-identity privacy, other attributes may require protection in different scenarios. For example, background environmental sounds (e.g., announcements at a train station) can reveal a speaker's location, necessitating anonymization that considers scenario-dependent audio cues alongside voice identity. 
Content privacy~\citep{williams2021revisiting,williams2022new,sinha2024safeguarding,hui2025securespeech} is also a major concern. 
Even if the latter is now attracted attention, most approaches treat named entities as the primary privacy-sensitive information and simply remove them --- a solution that is not always practical. An attacker may still infer sensitive meaning from surrounding context even if named entities are removed.

\textit{Reversibility and recovery requirements}:
To ensure strong privacy guarantees, most anonymization systems are designed to be irreversible. However, reversibility is required in certain use cases. For example, in forensic or security contexts, the original content and speaker-identity information should be recoverable to facilitate investigations. Recent work~\citep{zhang2025speechguard} explores recoverable anonymization methods enabling the selective restoration or user-customized modification of anonymized speech. This direction addresses forensic and authorized investigation scenarios where conditional decryption and voice recovery are necessary, while maintaining privacy under normal operation. A practical implementation could involve allowing users or authorized parties to maintain a secure mapping between original and anonymized speaker/content, enabling recovery when necessary.

\section{Conclusions}
\label{sec:conclusions}

The VoicePrivacy 2024 Challenge represents a significant milestone in voice anonymization research, demonstrating substantial progress in privacy-preserving speech technologies which simultaneously preserve linguistic and emotional content. 
The challenge attracted 36 submissions from 13 teams across 16 countries, representing both academic and non-academic organizations, significant international engagement, and substantial methodological diversity -- more than doubling participation compared to the previous challenge edition.
Many submitted systems evolved far beyond incremental baseline improvements, introducing state-of-the-art neural architectures combining neural audio codecs, attribute disentanglement, self-supervised learning models, and emotion-aware design.

Results show that strong privacy protection can often only be delivered using complex neural architectures and careful consideration of multiple competing objectives. 
Only six of the submitted systems achieved the highest privacy target of an EER $\geq$ 40\% while only four  met the full complement of other challenge requirements, illustrating the fundamental difficulty of protecting privacy while simultaneously preserving utility.
Results show that cascaded ASR+TTS systems achieve near-perfect anonymization but at the cost of emotion preservation.  Through multi-stage disentanglement and frame-level emotion distillation, neural codec-based approaches achieve a better balance between privacy and utility requirements.
Systems without explicit emotion-aware components mostly underperform those with dedicated emotion distillation, encoder integration, or emotion-matched pseudo-speaker selection. 

Five dominant technical trends emerged.  They include: 
the adoption of neural codecs providing 
frameworks for attribute disentanglement; 
the use of self-supervised learning and pretrained models for feature extraction; 
multi-stage attribute disentanglement of speaker identity, linguistic content, prosody, and emotion  
anonymization techniques which enable flexible privacy-utility trade-off through dynamic method selection;  
emotion-aware 
design using emotion encoders, distillation mechanisms and emotion-conditioned synthesis.

The best submitted systems significantly outperform baselines across all metrics.
Specifically, for individual metrics: 
EER improves from 34\% (\textbf{B5}) to 49.5\% (\textbf{T8-1}), UAR improves from 54\% (\textbf{B2}) to 65\% (\textbf{T10-1}), and WER reduces from 2.9\% (\textbf{B1}) to 2.37\% (\textbf{T9}).

Some challenges and other concerns persist. 
Other work~\cite{tomashenko2025first} reveals that
privacy evaluation is vulnerable to overestimation: 
four top-performing systems that initially achieved an EER $\geq$ 40\% were later shown to be vulnerable to attacks that reduced EERs to less that 20\%.
This finding emphasizes the importance of independent and continuous two-stage evaluation in which defenses and attacks are developed and revised by competing teams.
This challenge emphasizes the importance of independent or continuous two-stage evaluation frameworks in which anonymization systems and attackers are developed by different teams, 
ensuring that privacy assessments are based on independently verified threat models rather than single standardized attackers.

Emerging research directions include the study of multi-attribute and application-specific anonymization, real-time and computationally efficient solutions, and 
semantic-level
approaches leveraging prompt-based and LLM-driven methods for flexible, user-controllable anonymization. 
The evolution from simple baseline approaches to complex multi-stage disentanglement systems reflects rapid methodological development in voice anonymization research.
As the field continues to advance toward the next VoicePrivacy Challenge edition tentatively scheduled for 2026, addressing these evaluation limitations, expanding application scenarios, and ensuring fairness across diverse populations will be critical to developing robust, trustworthy, and widely deployable voice anonymization solutions.

\section*{Acknowledgment}

This work was supported by the French National Research Agency under the  SpeechPrivacy project (\url{https://anr.fr/Projet-ANR-23-CE23-0022}) and the IPoP project funded by the French Cybersecurity PEPR (\url{https://www.pepr-cybersecurite.fr/projet/ipop/}). This work was partially supported by JST, PRESTO Grant Number JPMJPR23P9, Japan. Experiments were carried out using the Grid’5000 testbed.
The challenge organizers thank \textit{Ünal Ege Gaznepoğlu} for his help with the code base.

\bibliographystyle{cas-model2-names}

\bibliography{cas-refs}

\appendix

\section{Training resources}\label{app:trainig_res}

Table~\ref{tab:data-models-final-list} presents the list of allowed training resources for anonymization systems.

\begin{longtable}[htbp!]{cl}
  \caption{List of models and data for training anonymization systems. $\star$  denotes multiple requests from different teams for the corresponding model/dataset.}\label{tab:data-models-final-list}\\
\toprule 
$ \cellcolor{gray!25} \textbf{\#}$	&	\cellcolor{gray!25}\textbf{Model}		\\ 
% \hhline{==}   
1	&	WavLM  Base~and~Large \citep{Chen2021WavLM} $\star\star\star\star$ \\		& \url{	https://github.com/microsoft/unilm/tree/master/wavlm	} \\ \midrule 
2	&	Whisper \citep{radford2023robust}	$\star\star$ \\ &	\url{	https://github.com/openai/whisper	} \\ \midrule
3	&	HuBERT \citep{hubert} $\star\star\star$ \\ &	 \url{	https://github.com/facebookresearch/fairseq/blob/main/examples/hubert	} \\ \midrule
	% &		 \url{		} \\
4	&	XLS-R \citep{babu2021xls}	$\star\star$ \\ & \url{	https://github.com/facebookresearch/fairseq/blob/main/examples/wav2vec/xlsr	} \\ \midrule
5	&	wav2vec 2.0 \citep{baevski2020wav2vec}	$\star\star\star$ \\ & \url{	https://github.com/facebookresearch/fairseq/tree/main/examples/wav2vec	} \\
	&		 \url{	https://dl.fbaipublicfiles.com/voxpopuli/models/wav2vec2_large_west_germanic_v2.pt	} \\ \midrule
6	&	wav2vec2-large-robust-12-ft-emotion-msp-dim \citep{wagner2023dawn}	 \\ & \url{https://huggingface.co/audeering/wav2vec2-large-robust-12-ft-emotion-msp-dim} \\ \midrule
7	&	ContentVec \citep{qian2022contentvec};	  \url{	https://github.com/auspicious3000/contentvec	} \\ \midrule
8	&	w2v-BERT \citep{chung2021w2v} \\ &	 \url{	https://github.com/facebookresearch/fairseq/tree/ust/examples/w2vbert	} \\ \midrule
9	&	ECAPA2 \citep{thienpondt2023ecapa2};  \url{	https://huggingface.co/Jenthe/ECAPA2	} \\ \midrule
10	&	ECAPA-TDNN	\citep{desplanques2020ecapa}; \url{	https://huggingface.co/speechbrain/spkrec-ecapa-voxceleb	} \\ \midrule
11	&	NaturalSpeech 3	\citep{ju2024naturalspeech}; \url{	https://huggingface.co/amphion/naturalspeech3_facodec	} \\ \midrule
12	&	\makecell[l]{NVIDIA~Hifi-GAN   Vocoder (en-US) \citep{kong2020hifi}};  \url{	https://huggingface.co/nvidia/tts_hifigan} \\ \midrule
13	&	CRDNN on CommonVoice 14.0 English \\ &	 \url{	https://huggingface.co/speechbrain/asr-crdnn-commonvoice-14-en	} \\ \midrule
14	&	Encodec \citep{encodec}; \url{	https://huggingface.co/facebook/encodec_24khz	} \\ \midrule
15	&	Bark; \url{	https://huggingface.co/suno/bark	}; 		 \url{	https://huggingface.co/erogol/bark/tree/main	} \\  
% \bottomrule
% \multicolumn{2}{l}{}  \\ 

\\
 \toprule 
$ \cellcolor{gray!25}\textbf{\#}$		&	\cellcolor{gray!25}\textbf{Dataset}		\\ 
% \hmidrule{==}

16	&	ESD \citep{zhou2021seen}	$\star\star\star\star$ \\ & \url{ 	https://hltsingapore.github.io/ESD/download.html} \\ \midrule
17	&	LibriSpeech \citep{panayotov2015librispeech}: train-clean-100, train-clean-360, train-other-500	$\star\star\star\star\star$\\ & \url{	https://www.openslr.org/12	} \\ \midrule
18	&	CREMA-D \citep{cao2014crema}	$\star\star$ \\ & \url{	https://github.com/CheyneyComputerScience/CREMA-D} \\ \midrule
19	&	RAVDESS \citep{livingstone2018ryerson}	 \\ & \url{	https://datasets.activeloop.ai/docs/ml/datasets/ravdess-dataset/} \\
	&		 \url{	https://zenodo.org/records/1188976} \\ \midrule
20	&	VCTK \citep{yamagishi2019cstr} $\star\star\star\star\star\star$ \\ &	 \url{	https://datashare.ed.ac.uk/handle/10283/2651};	 \url{	https://huggingface.co/datasets/vctk} \\ \midrule
21	&	SAVEE \citep{haq2009speaker}; \url{	http://kahlan.eps.surrey.ac.uk/savee/} \\
 & \url{https://www.kaggle.com/datasets/ejlok1/surrey-audiovisual-expressed-emotion-savee}	 \\ \midrule
22	&	EMO-DB	\citep{burkhardt2005database}; \url{ http://emodb.bilderbar.info/download/} \\ \midrule
23	&	LJSpeech \citep{ljspeech17}; \url{ https://keithito.com/LJ-Speech-Dataset/} \\ \midrule
24	&	Libri-light \citep{kahn2020libri} (only train part) $\star\star$	 \\ & \url{https://github.com/facebookresearch/libri-light/blob/main/data_preparation/README.md} \\ \midrule
25	&	VoxCeleb-1,2 \citep{chung2018voxceleb2} $\star\star\star$	\\ & \url{https://www.robots.ox.ac.uk/~vgg/data/voxceleb/index.html\#about} \\ \midrule
26	&	LibriTTS \citep{zen2019libritts}: train-clean-100,  train-clean-360, train-other-500	$\star\star\star\star$ \\ & \url{	https://openslr.org/60/	} \\ \midrule
27	&	CMU-MOSEI	\citep{zadeh2018multimodal};  \url{	http://multicomp.cs.cmu.edu/resources/cmu-mosei-dataset/	} \\ \midrule
28	&	MUSAN \citep{snyder2015musan}; \url{	https://www.openslr.org/17/	} \\ \midrule
29	&	RIR 	\citep{ko2017study}; \url{	https://www.openslr.org/28/	} \\ \midrule
30	&	VGAF \citep{sharma2021audio} (from EmotiW challenge); \url{	https://sites.google.com/view/emotiw2023} \\ 
& \url{https://www.kaggle.com/datasets/amirabdrahimov/vgaf-dataset} \\ \midrule
31	& MSP-Podcast \citep{lotfian2017building}  \\ & \url{https://ecs.utdallas.edu/research/researchlabs/msp-lab/MSP-Podcast.html} \\ 
% \bottomrule
%
\\
% \multicolumn{2}{l}{}  \\
\toprule 

\cellcolor{gray!25} $\textbf{\#}$		&	\cellcolor{gray!25}\textbf{Software with pretrained models}	 \\ 
% \hmidrule{==}
%
32	&	Resemblyzer:	 \url{	https://github.com/resemble-ai/Resemblyzer	}  $\star$\\
	&		 	Model: 	\url{https://github.com/resemble-ai/Resemblyzer/blob/master/resemblyzer/pretrained.pt	} \\ \midrule
33	&	VITS \citep{kim2021conditional}; \url{	https://github.com/jaywalnut310/vits/	} \\ 
	&			Models: \url{	 https://drive.google.com/drive/folders/1ksarh-cJf3F5eKJjLVWY0X1j1qsQqiS2	} \\ \midrule
34	&	PIPER pretrained on VITS:	 \url{	https://github.com/rhasspy/piper/?tab=readme-ov-file} \\ 
	 &	Models:  \url{	https://huggingface.co/datasets/rhasspy/piper-checkpoints/tree/main} \\ \midrule
35	&	RVC-Project: \url{https://github.com/RVC-Project} \\
	&			Models: \url{https://huggingface.co/lj1995/VoiceConversionWebUI/tree/main	} \\ \midrule
36	&	DISSC \citep{maimon2022speaking}; 	 \url{	  https://github.com/gallilmaimon/DISSC} \\ \bottomrule

\end{longtable}
\normalsize

\section{Results}\label{app:ranking}

Table~\ref{tab:results} shows performance metrics by system on development and test data.

\begin{figure}[h]
    \centering
\includegraphics[width=0.5\linewidth]{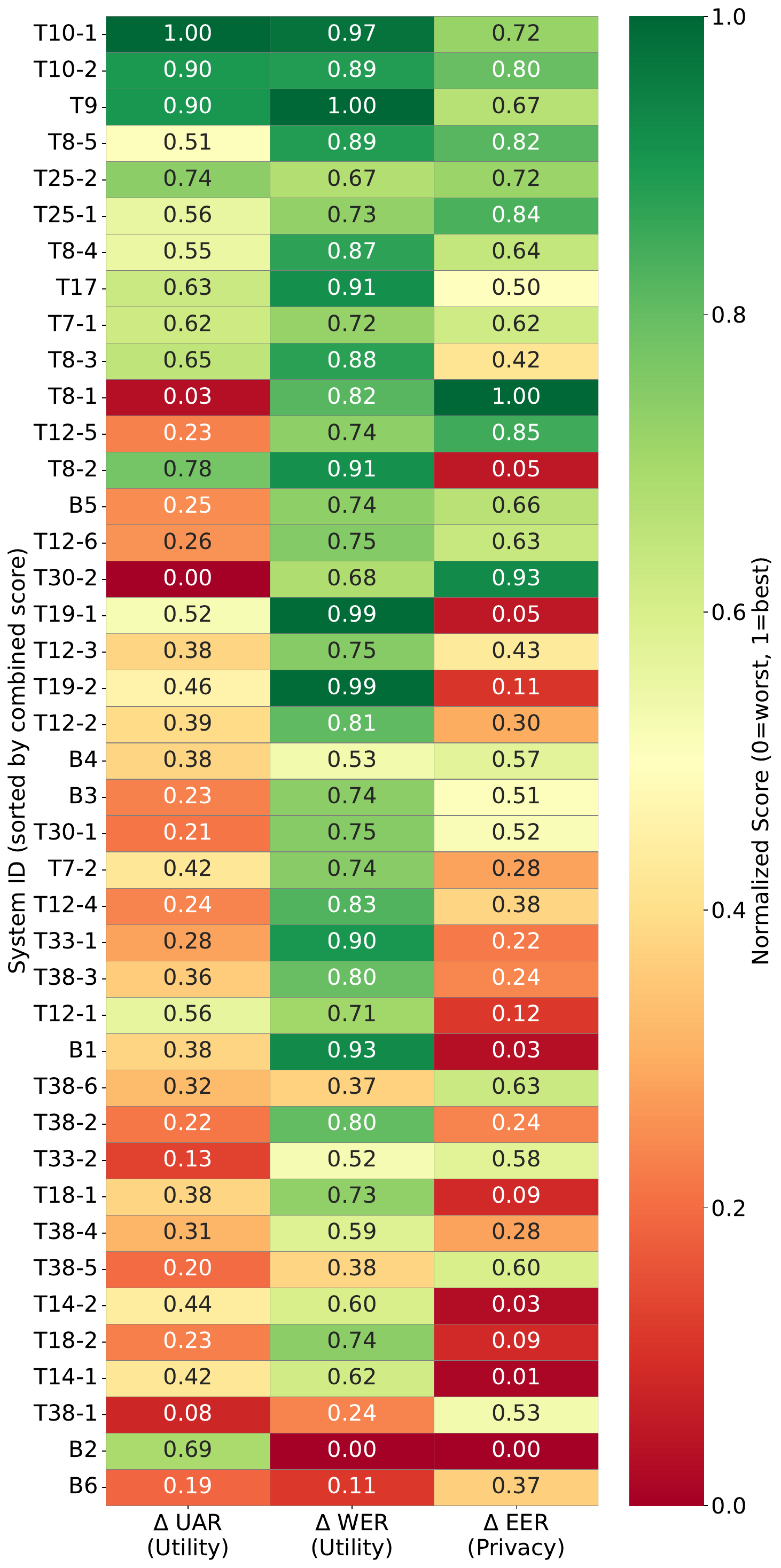}
    \caption{Normalized performance matrix showing all systems ranked by overall score (unofficial ranking).}
    \label{fig:heatmap-all}
\end{figure}

Figure~\ref{fig:heatmap-all} demonstrates 
normalized performance matrix showing all systems ranked by overall score (unofficial ranking). 
The heatmap presents a normalized performance comparison across multiple speech anonymization systems evaluated on three key metrics (relative change w.r.t to the results on the original data): $\Delta$UAR, $\Delta$WER, and $\Delta$EER. Values are normalized on a scale from 0 to 1, where 1 corresponds to the best performance and 0 to the worst. The $\Delta$UAR and $\Delta$WER metrics are inverted during normalization to reflect that lower values indicate better utility (less utility loss), while $\Delta$EER values are kept  as is to indicate higher privacy. 
Systems are ranked by a combined score, averaging their normalized scores across all three metrics, and the heatmap is sorted accordingly. Color coding from red to green visually highlights system strengths and weaknesses, with green indicating excellent performance and red indicating poorer performance.

\begin{table}[htbp]
\centering
\small
\caption{Performance metrics by system: EER,\% and WER.\% on \textit{LibriSpeech} and UAR,\% on \textit{IEMOCAP} development and test sets.}
\label{tab:results}
\begin{tabular}{lrrrrr|rrrrr}
\toprule
\multirow{3}{*}{\textbf{System ID}} 
& \multicolumn{5}{c}{\textbf{development}} 
& \multicolumn{5}{c}{\textbf{test}} \\
\cmidrule(lr){2-6} \cmidrule(lr){7-11}
& \multicolumn{3}{c}{\textbf{EER, \%}} 
& \textbf{UAR, \%} 
& \textbf{WER, \%} 
& \multicolumn{3}{c}{\textbf{EER, \%}} 
& \textbf{UAR, \%} 
& \textbf{WER, \%} \\
\cmidrule(lr){2-4} \cmidrule(lr){5-5} \cmidrule(lr){6-6} \cmidrule(lr){7-9} \cmidrule(lr){10-10} \cmidrule(lr){11-11}
& \textbf{female} & \textbf{male} & \textbf{aver} 
&  & 
& \textbf{female} & \textbf{male} & \textbf{aver} 
&  &  \\ \midrule
orig     & 10.51 & 0.93  & 5.72  & 69.08 & 1.80 & 8.76  & 0.42  & 4.59  & 71.10 & 1.85 \\ \midrule
B1       & 10.94 & 7.45  & 9.20  & 42.71 & 3.07 & 7.47  & 4.68  & 6.08  & 42.80 & 2.91 \\
B2       & 12.91 & 2.05  & 7.48  & 55.61 & 10.44& 7.48  & 1.56  & 4.52  & 53.50 & 9.95 \\
B3       & 28.43 & 22.04 & 25.24 & 38.09 & 4.29 & 27.92 & 26.72 & 27.32 & 37.60 & 4.35 \\
B4       & 34.37 & 31.06 & 32.72 & 41.97 & 6.15 & 29.37 & 31.16 & 30.27 & 42.80 & 5.90 \\
B5       & 35.82 & 32.92 & 34.37 & 38.08 & 4.73 & 33.95 & 34.73 & 34.34 & 38.20 & 4.37 \\
B6       & 25.14 & 20.96 & 23.05 & 36.39 & 9.69 & 21.15 & 21.14 & 21.15 & 36.10 & 9.09 \\
\midrule
T7-1     & 35.23 & 27.64 & 31.43 & 48.70 & 4.86 & 34.67 & 30.23 & 32.45 & 51.22 & 4.47 \\
T7-2     & 21.57 & 14.59 & 18.08 & 42.38 & 4.59 & 19.89 & 14.74 & 17.32 & 44.18 & 4.32 \\
T8-1     & 47.33 & 48.12 & 47.72 & 30.08 & 3.74 & 48.90 & 50.15 & 49.53 & 30.59 & 3.75 \\
T8-2     & 11.65 &  5.28 &  8.46 & 56.73 & 3.27 &  9.85 &  4.03 &  6.94 & 56.67 & 3.05 \\
T8-3     & 25.00 & 16.90 & 20.95 & 51.28 & 3.30 & 21.19 & 25.39 & 23.29 & 52.13 & 3.29 \\
T8-4     & 34.93 & 32.80 & 33.86 & 49.34 & 3.51 & 33.40 & 33.67 & 33.53 & 48.73 & 3.34 \\
T8-5     & 39.63 & 40.84 & 40.24 & 47.08 & 3.45 & 42.50 & 40.05 & 41.28 & 47.10 & 3.20 \\
T9       & 32.10 & 33.88 & 32.99 & 60.66 & 2.33 & 32.30 & 36.74 & 34.52 & 60.97 & 2.37 \\
T10-1    & 43.33 & 34.32 & 38.82 & 65.98 & 2.57 & 37.24 & 36.56 & 36.90 & 64.45 & 2.57 \\
T10-2    & 43.63 & 40.04 & 41.83 & 62.93 & 3.68 & 41.97 & 38.75 & 40.36 & 60.80 & 3.22 \\
T12-1    & 19.18 &  6.99 & 13.08 & 49.20 & 4.97 & 12.04 &  7.39 &  9.71 & 49.12 & 4.60 \\
T12-2    & 22.44 & 11.53 & 16.99 & 42.76 & 3.81 & 20.44 & 15.59 & 18.02 & 43.21 & 3.83 \\
T12-3    & 26.14 & 18.32 & 22.23 & 44.67 & 4.21 & 25.53 & 22.54 & 24.03 & 42.78 & 4.29 \\
T12-4    & 24.57 & 15.68 & 20.13 & 39.18 & 3.61 & 22.99 & 20.27 & 21.63 & 37.67 & 3.69 \\
T12-5    & 43.32 & 44.10 & 43.71 & 38.06 & 4.76 & 43.61 & 41.88 & 42.75 & 37.60 & 4.36 \\
T12-6    & 33.10 & 33.38 & 33.24 & 39.41 & 4.73 & 32.27 & 33.89 & 33.08 & 38.51 & 4.25 \\
T14-1    & 19.32 &  3.42 & 11.37 & 45.45 & 6.14 &  7.12 &  3.16 &  5.14 & 44.23 & 5.28 \\
T14-2    & 17.33 &  2.52 &  9.92 & 45.57 & 6.59 &  8.96 &  2.67 &  5.81 & 44.85 & 5.40 \\
T17      & 36.08 & 41.44 & 38.76 & 51.71 & 2.95 & 34.13 & 20.26 & 27.20 & 51.41 & 3.02 \\
T18-1    & 14.77 &  5.28 & 10.03 & 45.09 & 4.46 & 10.95 &  6.01 &  8.48 & 42.74 & 4.39 \\
T18-2    & 13.78 &  4.81 &  9.29 & 43.52 & 4.79 & 10.77 &  6.04 &  8.40 & 37.37 & 4.34 \\
T19-1    & 11.79 &  4.33 &  8.06 & 46.94 & 2.60 &  8.39 &  5.57 &  6.98 & 47.66 & 2.45 \\
T19-2    & 16.62 &  7.01 & 11.82 & 43.20 & 2.48 & 11.11 &  7.76 &  9.43 & 45.50 & 2.44 \\
T25-1    & 42.65 & 40.06 & 41.35 & 52.38 & 4.73 & 42.34 & 41.91 & 42.13 & 48.99 & 4.44 \\
T25-2    & 36.65 & 35.72 & 36.18 & 55.31 & 5.19 & 37.74 & 35.85 & 36.79 & 55.39 & 4.85 \\
T30-2    & 45.59 & 48.45 & 47.02 & 28.45 & 5.02 & 45.65 & 47.00 & 46.33 & 29.44 & 4.78 \\
T30-1    & 29.10 & 22.20 & 25.65 & 38.48 & 4.13 & 30.11 & 25.39 & 27.75 & 36.78 & 4.28 \\
T33-2    & 29.71 & 36.80 & 33.25 & 37.31 & 6.17 & 23.89 & 37.01 & 30.45 & 34.00 & 5.98 \\
T33-1    & 22.16 & 13.35 & 17.76 & 38.69 & 3.16 & 14.23 & 14.69 & 14.46 & 39.27 & 3.12 \\
T38-1    & 40.31 & 34.77 & 37.54 & 32.30 & 8.47 & 27.73 & 29.18 & 28.45 & 32.17 & 8.16 \\
T38-2    & 25.43 & 15.84 & 20.63 & 37.86 & 4.18 & 13.51 & 16.97 & 15.24 & 36.98 & 3.86 \\
T38-3    & 25.00 & 16.61 & 20.81 & 44.51 & 4.20 & 15.15 & 15.60 & 15.37 & 42.08 & 3.91 \\
T38-4    & 27.70 & 21.90 & 24.80 & 42.51 & 5.85 & 16.03 & 18.48 & 17.26 & 40.45 & 5.51 \\
T38-5    & 41.19 & 37.44 & 39.31 & 36.08 & 7.78 & 31.02 & 31.85 & 31.44 & 36.37 & 7.07 \\
T38-6    & 42.58 & 38.35 & 40.47 & 41.19 & 7.80 & 31.91 & 33.63 & 32.77 & 40.79 & 7.11 \\ \midrule
% post-eval:
\cellcolor{Gray!5}{T25-p1}	 & \cellcolor{Gray!5}45.12	 & \cellcolor{Gray!5}41.22  &	\cellcolor{Gray!5}43.17  &	\cellcolor{Gray!5}53.84	 & \cellcolor{Gray!5}4.72  &	\cellcolor{Gray!5}44.19  &	\cellcolor{Gray!5}41.40  &	\cellcolor{Gray!5}42.80  &	\cellcolor{Gray!5}50.07	 & \cellcolor{Gray!5}4.45 \\ 
\cellcolor{Gray!5}{T25-p2}  &	\cellcolor{Gray!5}38.29  &	\cellcolor{Gray!5}34.11  &	\cellcolor{Gray!5}36.20	 & \cellcolor{Gray!5}57.86  &	\cellcolor{Gray!5}4.98	 & \cellcolor{Gray!5}37.94	 & \cellcolor{Gray!5}35.51  &	\cellcolor{Gray!5}36.73  &	\cellcolor{Gray!5}55.79 &	\cellcolor{Gray!5}5.12\\
\bottomrule
\end{tabular}
\end{table}

\end{document}